\def\cvprPaperID{11727}
\def\confName{ICCV}
\def\confYear{2023}

\def\paperTitle{RefineVIS: Video Instance Segmentation with Temporal Attention Refinement}

\def\authorBlock{
    Andre Abrantes \qquad
    Jiang Wang \qquad
    Peng Chu \qquad
    Quanzeng You \qquad
    Zicheng Liu \\
    Microsoft Cloud and AI \\
    {\tt\small \{abrantes, jiangwang, pengchu, qyou, zliu\}@microsoft.com}
}

\newif\ifreview 
\newif\ifarxiv \newcommand{\arxiv}{\arxivtrue}
\newif\ifcamera 
\newif\ifrebuttal

\arxiv 

\pdfoutput=1
\documentclass[10pt,twocolumn,letterpaper]{article}

\ifreview \usepackage[review]{cvpr} \fi
\ifarxiv \usepackage[pagenumbers]{cvpr} \fi
\ifrebuttal \usepackage[rebuttal]{cvpr} \fi
\ifcamera \usepackage{cvpr} \fi

\usepackage{graphicx}
\usepackage{amsmath}
\usepackage{amssymb}
\usepackage{booktabs}

\usepackage{times}
\usepackage{microtype}
\usepackage{epsfig}
\usepackage[table,xcdraw]{xcolor}
\usepackage{caption}
\usepackage{float}
\usepackage{placeins}
\usepackage{color, colortbl}
\usepackage{stfloats}
\usepackage{enumitem}
\usepackage{tabularx}
\usepackage{xstring}
\usepackage{multirow}
\usepackage{xspace}
\usepackage{url}
\usepackage{subcaption}
\usepackage{xcolor}
\usepackage[hang,flushmargin]{footmisc}

\ifcamera \usepackage[accsupp]{axessibility} \fi

% ...

\ifarxiv  \fi

\newcommand{\R}[1]{{%
    \textbf{%
        \ifstrequal{#1}{1}{\textcolor{red}{R#1}}{%
        \ifstrequal{#1}{2}{\textcolor{blue}{R#1}}{%
        \ifstrequal{#1}{3}{\textcolor{magenta}{R#1}}{%
        \ifstrequal{#1}{4}{\textcolor{teal}{R#1}}{%
                           \textcolor{cyan}{R#1}%
        }}}}%
    }%
}}

\usepackage{xr-hyper}

\makeatletter
\newcommand*{\addFileDependency}[1]{
  \typeout{(#1)}
  \@addtofilelist{#1}
  \IfFileExists{#1}{}{\typeout{No file #1.}}
}

\makeatother

\usepackage[pagebackref,breaklinks,colorlinks]{hyperref}
\usepackage[capitalize]{cleveref}
\crefname{section}{Sec.}{Secs.}
\crefname{table}{Table}{Tables}
\crefname{figure}{Fig.}{Figs.}

\begin{document}
\title{\paperTitle}
\author{\authorBlock}
\maketitle

\begin{abstract}
We introduce a novel framework called RefineVIS for Video Instance Segmentation (VIS) that achieves good object association between frames and accurate segmentation masks by iteratively refining the representations using sequence context. RefineVIS learns two separate representations on top of an off-the-shelf frame-level image instance segmentation model: an association representation responsible for associating objects across frames and a segmentation representation that produces accurate segmentation masks. Contrastive learning is utilized to learn temporally stable association representations. A Temporal Attention Refinement (TAR) module learns discriminative segmentation representations by exploiting temporal relationships and a novel temporal contrastive denoising technique. Our method supports both online and offline inference. It achieves state-of-the-art video instance segmentation accuracy on YouTube-VIS 2019 (64.4 AP), Youtube-VIS 2021 (61.4 AP), and OVIS (46.1 AP) datasets. The visualization shows that the TAR module can generate more accurate instance segmentation masks, particularly for challenging cases such as highly occluded objects.


\end{abstract}
\section{Introduction}
\label{sec:intro}

Video Instance Segmentation (VIS)~\cite{yang2019video} is a computer vision task that follows multiple objects of different categories throughout a video while predicting the pixels each one occupies at each frame. To achieve good video instance segmentation accuracy, it is crucial to utilize temporal understanding in addition to modeling the objects' appearance on each frame because an object can have a rapidly varying appearance or become almost invisible in some frames due to occlusion, motion blur, or small size. Modeling objects' appearance independently is difficult for these challenging scenarios.

Recent works~\cite{cheng2021mask2formervideo, wu2021seqformer, heo2022vita} showed promising results on using Transformers to model the spatial and temporal relationship for VIS. However, these methods usually do not have very good results in datasets with complex videos, such as OVIS~\cite{qi2022occluded}, because of the challenges of associating an object with varying appearances in videos that are long and filled with occlusion. While recent works on online methods~\cite{IDOL} can achieve good association results, they do not model the object's temporal relationship for predicting segmentation masks.

A key observation in this paper is that we need to learn separate representations for two critical temporal understandings in VIS: associating objects across frames and producing accurate segmentation masks for each frame. An association representation should be stable across multiple frames. In contrast, a segmentation representation should characterize the object appearance at each frame by considering the corresponding object appearance of the adjacent frames. Previous works mainly improve the association representation \cite{heo2022vita,IDOL}. However, these methods either do not use temporal information to improve segmentation or use a single representation for both tasks, which usually leads to sub-optimal video segmentation mask quality.

\begin{figure*}
    \centering
    \subfloat{\includegraphics[width=0.475\textwidth]{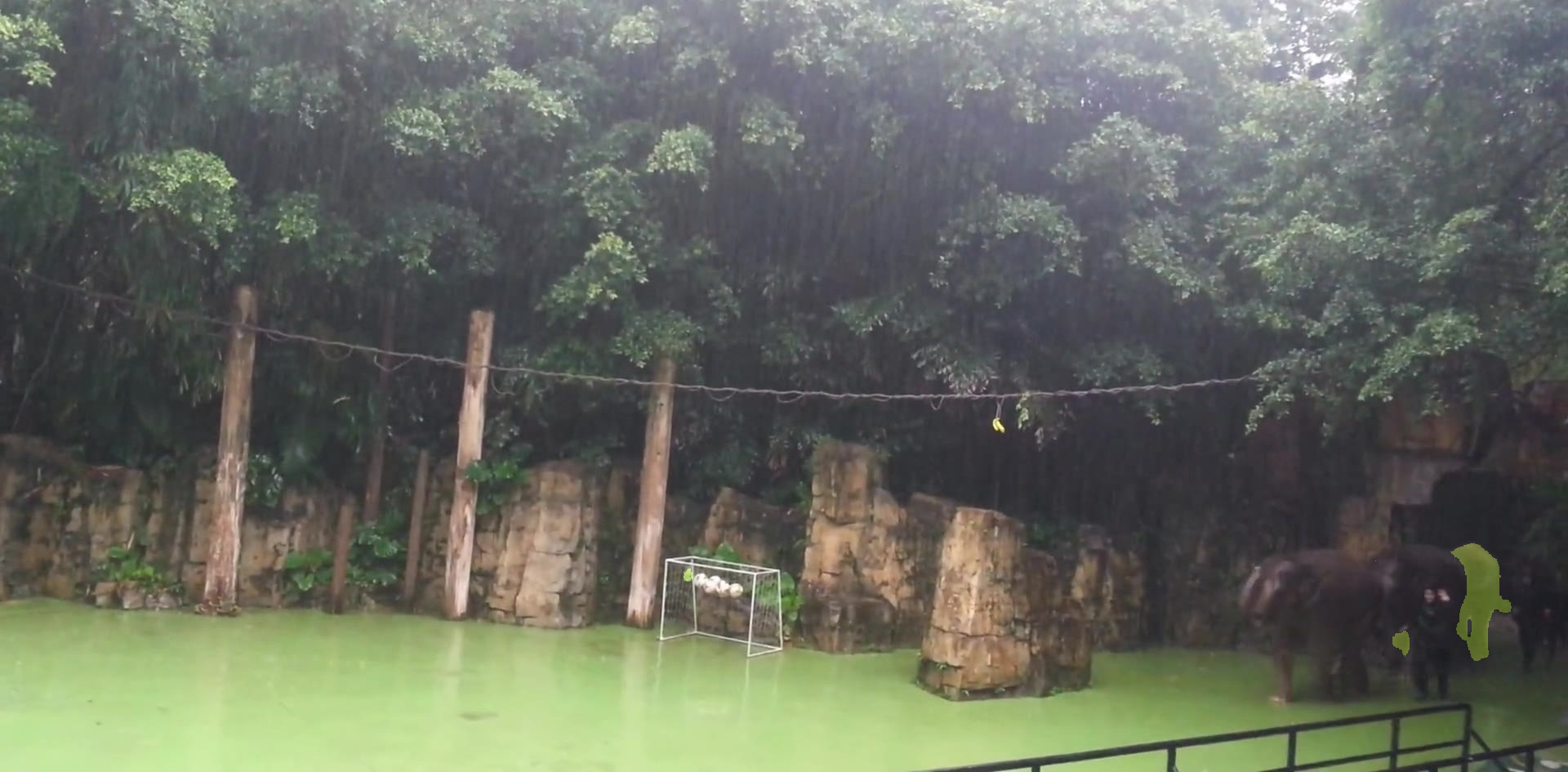}}
    \hfill
    \subfloat{\includegraphics[width=0.475\textwidth]{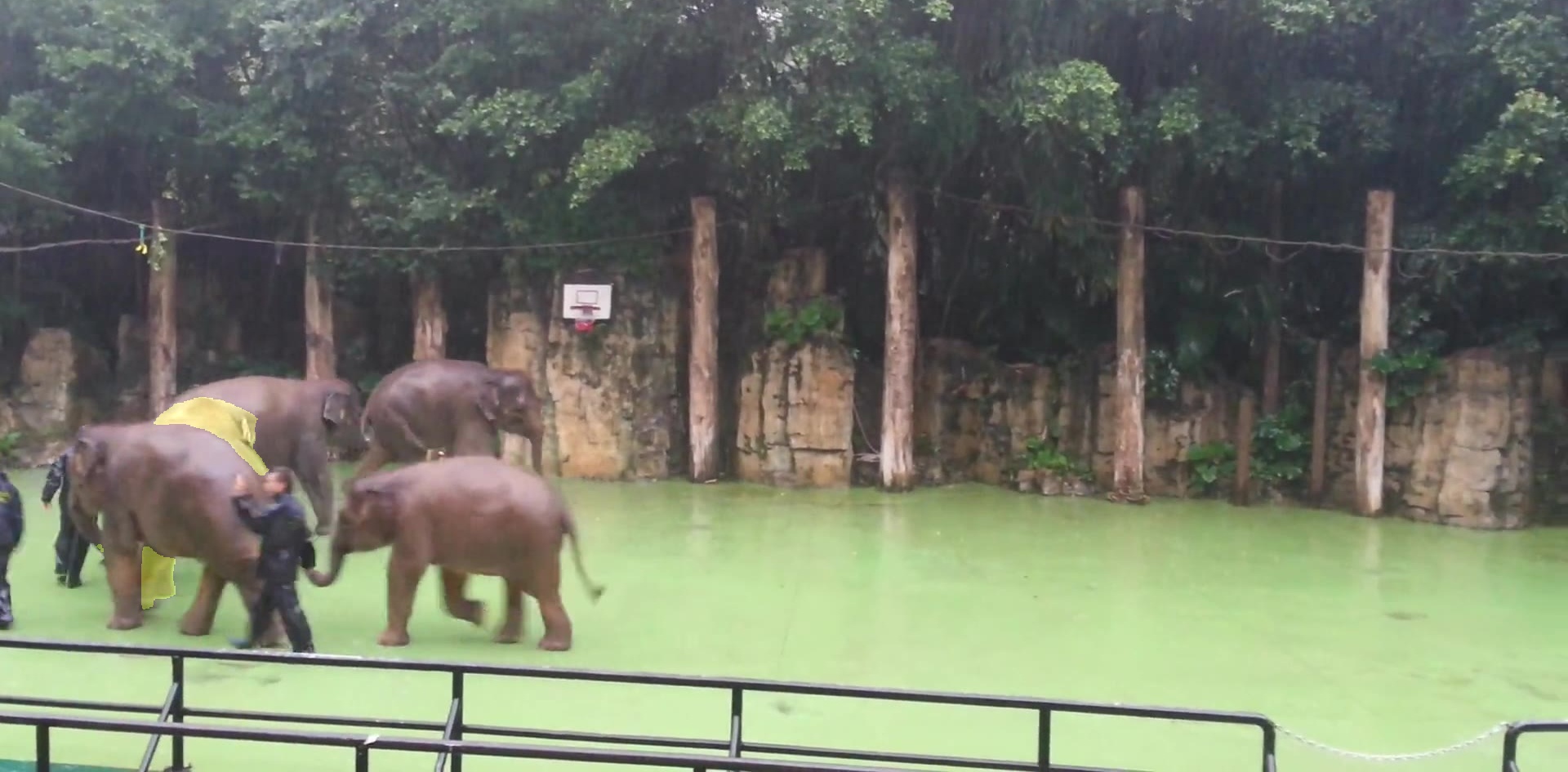}}
    \caption{
        Selected frames from an OVIS clip showing the high occlusion suffered by the third elephant (yellow mask) in two different frames.
        Our refinement layer was able to recover both masks even though the frame level model (detector + tracker) had lost the animal both times.
    }
    \label{fig:image_examples}
\end{figure*}

Based on this observation, we propose a novel VIS framework called RefineVIS that learns the association and segmentation representations separately. RefineVIS is built upon image-level instance segmentation models because they are simpler to enable online inference and take advantage of the advances in image-level instance segmentation. An image-level instance segmentation model outputs the object class, bounding box, and segmentation mask and an instance embedding for each instance in one frame. An association representation is learned on top of instance embeddings using contrastive learning similar to IDOL~\cite{IDOL}. After the association, a temporal module leveraging the novel Temporal Attention Refinement (TAR) layer is proposed to learn segmentation representation that utilizes the temporal relationship of the instance appearance to predict more accurate segmentation masks. A novel temporal contrastive denoising technique is proposed to improve the training of the temporal module. We find that TAR is crucial for predicting accurate segmentation masks when objects are severely occluded or on ambiguous angles. Figure \ref{fig:image_examples} is an example of severe occlusion, where the proposed method can predict accurate segmentation masks. In contrast, the baseline method cannot find the object at all. The segmentation representation is also employed to predict the instance classification categories.

The modules in our framework are learned end-to-end, and the framework supports both online and offline inference.
Experiments on YouTube-VIS 2019, Youtube-VIS 2021, and OVIS datasets demonstrate that our method achieves state-of-the-art results on all three benchmark datasets. Ablation studies show clear advantages of applying the TAR module for image-level instance segmentation models with different backbones.
Moreover, the proposed method predicts more accurate segmentation mask predictions in our visualization, particularly for challenging cases.

\section{Related Work}
\label{sec:related}
\textbf{Offline Video Instance Segmentation} is designed to predict the instance masks of a video clip at once. Most offline methods utilize an image feature encoder to encode the frames into frame-level image embeddings and build a spatiotemporal decoder that outputs the masks and class labels for each instance trajectory. VisTR~\cite{wang2021end} adopts Transformer~\cite{vaswani2017attention} in the decoder with instance queries for the whole clip. IFC~\cite{hwang2021video} proposes an inter-frame communication transformer separating spatial and temporal attention to reduce computation and memory usage significantly. Similarly, Seqformer~\cite{wu2021seqformer} creates a communication mechanism between frames using instance queries.
TeViT~\cite{yang2022temporally} introduces a new backbone to facilitate spatiotemporal feature extraction and improve VIS results.
Mask2Former~\cite{cheng2021mask2formervideo} substantially improves the performance of offline video instance segmentation methods with a mask attention mechanism.
More recently, VITA~\cite{heo2022vita} sets the new state-of-the-art in offline VIS by modeling the video as a set of object tokens. It also uses a similarity loss to learn better associations.

In general, offline methods achieve better accuracy because they can use information from past and future frames together. However, they are more computationally expensive and consume more memory. In addition, since it is impossible to fit the whole video into memory for long videos, offline video instance segmentation models usually split a long video into clips, process each clip independently, and connect the clip predictions in a post-processing step. This process is cumbersome and usually leads to low performance on long videos. As a result, supporting online inference is preferred.

\textbf{Online Video Instance Segmentation}. Most online methods are built upon image-level instance segmentation with an additional tracking mechanism to associate instances across videos. MaskTrack R-CNN~\cite{yang2019video} extends Mask R-CNN~\cite{he2017mask}. It uses a tracking head that associates instances by leveraging multiple cues such as instance appearance embedding similarity, semantic consistency, spatial coherence, and detection confidence.
Follow-up methods~\cite{cao2020sipmask,liu2021sg,yang2021crossover,li2021spatial} choose similar architectures and perform association at later stages.
Others introduced mechanisms for the temporal propagation of object information \cite{han2022visolo,ke2021prototypical}.
Online methods tend to be computationally efficient, which is especially important when handling long videos.
However, despite the desirable properties, the accuracy of online instance segmentation models is usually lower than offline instance segmentation methods.

Recently, IDOL~\cite{IDOL} introduced a powerful contrastive learning-based loss to learn more discriminative association representation with better consistency, leading to significantly more accurate associations. IDOL can perform better video instance segmentation than offline methods, particularly on datasets containing challenging videos such as OVIS~\cite{qi2022occluded}. However, as segmentation masks are predicted independently for each frame, it is difficult for IDOL to predict accurate segmentation masks for complex cases, such as severe occlusion or ambiguous angles, without leveraging temporal information.

\textbf{Temporal Attention in VIS}. Multiple recent works have employed attention layers to VIS.
MS-STS \cite{mssts} introduced a multi-scale spatio-temporal attention module and a new loss for better foreground-background separability.
InsPro \cite{inspro} presented a new deduplication loss and an inter-query attention mechanism for enriching queries.
TAFormer \cite{taformer} introduced a spatio-temporal, multi-scale attention module in the transformer encoder, a temporal self-attention module in the decoder, and an instance-level contrastive loss.
IFR \cite{ifr2023} employed a strong recurrent spatio-temporal attention mechanism and a new discriminative loss for a more consistent association.
They all share the characteristic of mixing spatial and temporal attention within each of their encoder and/or decoder layers.
Additionally, they use embedding or query propagation across frames for object association.

Different from these methods, we propose a novel Temporal Attention Refinement (TAR) module that operates separately from the spatial layers and serves a different role in our framework, RefineVIS.
TAR can predict accurate segmentation masks for challenging cases by learning a representation that models the temporal relationship.
By delegating spatial information extraction to a frame-level module, TAR can focus on the temporal aspect of VIS and improve object embeddings.
This independence allows our model to work in either online or offline mode.

RefineVIS was designed for easy substitution of each component with any off-the-shelf model.
In addition, we introduce novel temporal Contrastive Denoising (CDN) extensions for training RefineVIS.
Together, these components provide the temporal information that image-level models lack, achieving SOTA performance on multiple VIS benchmarks without the need to couple spatial and temporal processing.

\section{Method}
\label{sec:method}

In Video Instance Segmentation (VIS), a series of $N$ objects in a video composed of $T$ frames must have their segmentation masks and classes predicted.

A VIS model will output a series of segmentation masks $m_i^t$ at each frame $t$ and a classification label $c_i$ for the $i$-th object.
If an object $i$ does not exist at a frame $t$, its corresponding $m_i^t$ must be empty.

\subsection{Overview of RefineVIS}
At a high level, our model is composed of three stages:
\begin{enumerate}
  \item \textbf{A frame-level module} extracts spatial bounding boxes, masks, and classification labels from each individual input frame.
  \item \textbf{An association module} matches object predictions across frames into tracklets.
  \item \textbf{A temporal refinement module}, powered by a Temporal Attention Refinement (TAR) layer, that improves predictions by spreading temporal information within each tracklet.
\end{enumerate}

\begin{figure*}[!htbp]
    \centering
    \includegraphics[width=\textwidth]{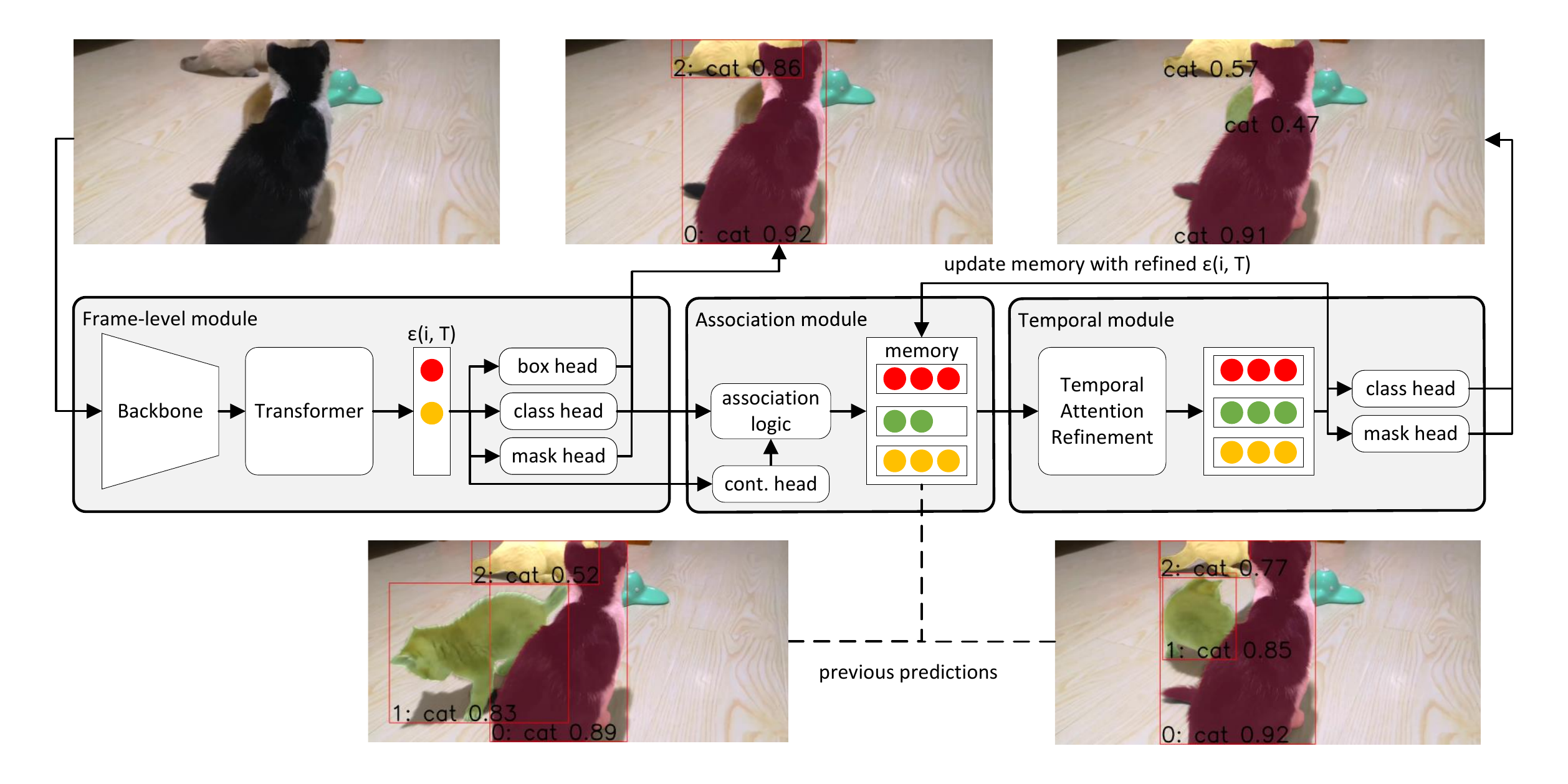}
    \caption{
    Overall architecture of RefineVIS showing the three stages and how they communicate.
    A clip with three cats is being processed.
    The first row shows the current frame being processed as well as its partial output after the frame-level predictions and its final output after refinement.
    In the second row, the three modules of RefineVIS are shown: frame-level, association and temporal.
    The last row shows the predictions from previous frames contained in the association memory.
    Notice how the cat with green mask was lost by the frame-level predictor but was recovered by refinement.
    Also notice how the cat with red mask had its mask improved and the tip of its tail recovered.
    The circles represent the objects' embeddings and the colors associate them with the segmented cats.
    }
    \label{fig:architecture}
\end{figure*}

Figure \ref{fig:architecture} illustrates the RefineVIS architecture.
The frame-level module predicts frame-level bounding boxes, masks, and classification labels. 
It also outputs the object embedding for each object.
The association module learns stable association representations from object embeddings and combines them with the frame-level predictions to associate frame-level objects into tracklets.
Finally, the refinement module uses the tracklet association and object embeddings to produce refined, temporal-aware predictions.

One benefit of the three-stage design is that we can easily upgrade the models of each stage to incorporate the recent development of image instance segmentation or association algorithms while keeping the other parts unchanged.

Two inference modes are possible in our model: offline and online.
In offline inference, the model will output the prediction for the whole sequence at the same time.
In online inference, the model will take one frame at a time and output the predictions for the frames sequentially.

Our model is trained end to end by combining the losses from the three stages.
\begin{equation}
	\mathcal{L} = \mathcal{L}_\text{frame} + \mathcal{L}_\text{association} + \mathcal{L}_\text{refine}
\end{equation}

\subsection{Frame-Level Module}
\label{subsec:frame-level}

In our architecture, the frame-level module extracts each frame's spatial bounding boxes, masks, and classification labels.
We use a DETR-like model as our frame-level module because it is the state-of-the-art image instance segmentation model, but any image instance segmentation model should work.

Given an input video $x_v \in \mathbb{R} ^{T \times 3 \times H_{0} \times W_{0}}$ with 3 color channels and $T$ frames of resolution $H \times W$, a backbone will extract one or multiple visual feature maps $\{f_t\}_{t=1}^{T} \in \mathbb{R}^{C \times H \times W}$ from $x_v$.
$C$ is usually a number as 2048 and indicates a deep feature representation of the input images, and, in contrast, $H$ and $W$ are typically lower resolution than $H_0$ and $W_0$.
In general, $f_t$ is extracted independently for each frame $t$.

The pair transformer encoder/decoder will consume the backbone features $\{f_t\}_{t=1}^{T}$ and will consecutively apply multiple layers of Multi-head Self-Attention for a set of queries over image features.
They produce a series of $T \times N_q$ per-frame, per-object representations called embeddings: $\mathcal{E}_{\text{obj}} \in \mathbb{R}^{T \times N_q \times C_{\mathcal{E}}}$.
$C_{\mathcal{E}}$ is the embedding dimension, and a typical value for it is 256. $N_q$ is the number of queries, which usually ranges from 100 to 300.
These queries are usually learnable, and they search for individual objects in the intermediary feature maps to create query-specific embeddings.
Prediction heads will consume these embeddings and generate predictions.
A mask head will predict segmentation masks of individual objects, a detection head will predict bounding boxes over objects, and a classification head will predict their class labels.

In a typical design, the classification head uses fully connected layers and sigmoid activation to predict the class probability.
The detection head regresses four encoded corners of the detection box. The mask head predicts a query-specific convolution kernel and applies it on the feature map to predict masks.

In a typical DETR model, the frame-level loss is composed of:
\begin{equation}
\label{eq:detr_loss}
	\mathcal{L}_\text{frame} = \mathcal{L}_\text{cls} + \mathcal{L}_\text{box} + \mathcal{L}_\text{mask}
\end{equation}

Following the typical DETR-like model design, $\mathcal{L}_\text{cls}$ is the focal loss, $\mathcal{L}_\text{box}$ is made of L1 and GIOU, and $\mathcal{L}_\text{mask}$ contains a cross-entropy and a DICE components.

\subsection{Object Association Module}

The object association module connects frame-level objects from multiple frames into tracklets. 
In this paper, we learn a stable association representation using a contrastive learning loss.
In contrastive learning, we always sample two frames from the video; the first is called the key frame, and the other is called the reference frame.
Objects in key and reference frames are grouped into positive and negative association pairs based on ground-truth matchings.
Their regular object embeddings are projected into contrastive embeddings by a contrastive head and used for association.
The contrastive association loss can be defined as:
\begin{equation}
    \mathcal{L}_\text{association} = \log[1 + \sum_{k^{+}}{\sum_{k^{-}}{\exp(v \cdot k^{-} - v \cdot k^{+})}}]
\end{equation}

Where $v$ are contrastive embeddings from each object in a key frame and $k^{+}$ and $k^{-}$ are the associated sets of positive and negative contrastive embeddings from the other frame.

After learning the stable association algorithm, the association algorithm connects the frame-level objects into tracklets using the association representation, leveraging the frame-level predictions.
In this paper, we employ the association algorithm in IDOL~\cite{IDOL}, an online association algorithm.
Our association module always runs in online mode during inference.

\subsection{Temporal Refinement Module}

Once objects from multiple frames are predicted and associated into tracklets, our refinement process works within each tracklet to refine segmentation masks and classes.
The temporal refinement module's inputs are the object embeddings of the objects within a tracklet from the frame-level model.


The refinement process consists of a Multi-head Self-Attention (MSA) layer followed by a Feed-Forward Network (FFN) layer.
The self-attention layer looks at all objects within a tracklet and propagates information among them.
We encode the absolute frame sequence number and add temporal positional encodings to the object embeddings to make the self-attention temporally location-aware.

Let $\mathcal{E}^t_i$ denote the object embedding from tracklet $i$ at frame $t$ and $\hat{\mathcal{E}^t_i}$ its refined version.
At a given frame $T$, a self-attention operation is computed within each tracklet $i$ comprising a temporal window of duration $W$ frames.
$\mathcal{E}^T_i$ is refined according to:
\begin{equation}
    \begin{split}
    \hat{\mathcal{E}^{T}_{i}} = \text{FFN}(\text{MSA}(\text{Concat}(\mathcal{E}^{T}_{i}, \{\hat{\mathcal{E}}^{t}_{i}\}^{T-1}_{t = T - W + 1}))), \\
    i=1, ..., N
    \end{split}
\end{equation}

With this temporal attention mechanism, TAR can refine class and mask predictions from the frame-level model.
In particular, TAR can improve and even recover masks when objects are erroneously not predicted by the frame-level model.
In both training and inference, when a tracklet does not contain a matching in a specific frame, we use a zero object embedding as the input to the TAR module, which helps TAR to learn to employ embeddings from nearby frames to make accurate predictions.

The losses for the temporal refinement module is: 
\begin{equation}
\label{eq:refine}
	\mathcal{L}_\text{refine} = \mathcal{L}_\text{cls} + \mathcal{L}_\text{mask}
\end{equation}

Where the loss components follow the same as in Equation \ref{eq:detr_loss} but handle predictions and ground truths from both input frames.

\subsection{Temporal Constrastive Denoising}

The Contrastive Denoising (CDN) technique was introduced by DINO~\cite{zhang2022dino} and MaskDINO~\cite{li2022maskdino}.
CDN queries are extra queries used in transformer decoders.
The purpose of CDN queries is to provide additional training signals, and they are not utilized in inference.
During training, a set of additional positive and negative queries are generated from the ground-truth bounding boxes and classes by applying different noise levels.
Positive queries are generated by applying a small amount of noise and are expected to reconstruct their corresponding ground-truth object classes, bounding boxes, and masks.
Negative queries are generated by applying higher noise levels and are expected to predict ``no object."
The standard classification, detection, and mask losses are applied using the corresponding expected ground truth, 
CDN queries are beneficial when only limited training data is available.

We propose two novel CDN queries to train our RefineVIS framework: Association CDN and Temporal CDN.
With \textbf{Contrastive CDN}, we improve contrastive learning in our association module.
Multiple CDN groups are created for each frame, each containing one positive and one negative CDN query for each ground truth object.
CDN contrastive pairs are made by associating the positive CDN queries in the key frame with other CDN queries from the reference frame.
For positive contrastive pairs, they are associated with all the positive CDN queries from the same tracklet in the reference frame.
For negative contrastive pairs, they are associated with the positive and negative CDN queries from the other tracklets in the reference frame.
A positive CDN query is not associated with its negative counterpart, as they might still be significantly similar.

For \textbf{Temporal CDN}, positive and negative CDN queries from the two frames are grouped into tracklets and used as inputs to the temporal refinement module.
We group the positive and negative frame-level queries into positive and negative tracklets based on their ground truth association.
Positive tracklets are expected to predict the corresponding masks and classification labels, while negative tracklets are expected to predict empty masks and class scores with zero confidence.

\subsection{Inference Mode}

Temporal refinement module can operate in both online and offline modes during inference.

In online mode, TAR keeps a memory window of size $W$, which contains the object embeddings from the current and past frames grouped into tracklets.
The TAR module runs self-attention on all the object embeddings in the memory but only outputs the refined predictions for the current frame.
After refinement, we update the object embeddings of the current frame to carry this refined information to the next iteration.
For applications such as streaming processing, online inference is required.

In offline mode, the object embeddings are split into temporal clips.
Clips are processed independently to predict the outputs for the whole clip at the same time.
The offline mode can leverage past and future embeddings when refining objects at a specific time and generally performs better.

\section{Experiments}
\label{sec:experiments}

We evaluate our model on YouTubeVIS-2019~\cite{yang2019video}, YouTubeVIS-2021~\cite{youtubevis2021}, and OVIS\cite{qi2022occluded} datasets.
YouTubeVIS-2019 is a collection of short videos (up to 36 frames) with segmented objects from 40 classes.
YouTubeVIS-2021 extends its predecessor dataset with more videos, improved categories, and doubles the number of annotations. It also introduces longer videos of up to 84 frames.
OVIS is a more challenging dataset because it contains short and long videos (from 15 to 292 frames) and 25 classes. Its objects are frequently severely occluded and in crowded scenes.

The models are evaluated using the regular validation subsets on the official evaluation servers of their respective datasets. For the YoutubeVIS-2019/2021 and OVIS datasets, we report our model's mean average precision (AP) and mean average recall (AR) over all classes. We also report AP measured at $0.50$ and $0.75$ IOU thresholds as $\text{AP}_{50}$ and $\text{AP}_{75}$ respectively.

\subsection{Implementation Details}
\label{sec:build_model}

For the frame-level module, most experiments are performed on our own implementation of MaskDINO~\cite{li2022maskdino} by extending the official DINO detection implementation~\cite{zhang2022dino}. We generally find using MaskDINO as our frame-level module achieves the best segmentation accuracy in our experiments. The structure of the MaskDINO is largely retained, including its class, box, mask heads, and regular CDN. The weights for the classification and mask heads are shared with the corresponding heads in the temporal refinement module because we find that sharing weights between these two modules leads to better accuracy.



During training, the association algorithm uses a lower threshold of IoU and prediction confidence so that the training gets sufficient supervision signals, particularly at the beginning of training when the predictions are noisy.

Our model is trained with both ResNet-50 and Swin-L backbones. We use NVIDIA V-100 GPUs to train and evaluate our ResNet-50 models and A-100 GPUs for our Swin-L models. Batch sizes of 16 clips are used to train models with both types of backbones.


Our training schedule contains three phases: pretraining with MS-COCO images, pretraining with MS-COCO pseudo videos, and training with VIS clips. We first pre-train our image-level model on the MS-COCO 2017 dataset \cite{mscoco} for detection and segmentation following a standard image instance segmentation training procedure.
Next, we create pseudo videos from MS-COCO images as other recent works~\cite{IDOL,heo2022vita,wu2021seqformer} and pre-train the whole RefineVIS model for 30k iterations. 
Finally, we train the model on the target VIS dataset.

In both video pretraining and VIS training, we randomly sample two frames from the sequence that are at most 10 frames apart.
We use both frames to train our association and TAR modules but only compute the image-level losses on the first frame.
The training signals from the association and TAR modules still back-propagate to the image-level models in both frames.
We find this setup leads to more stable training empirically.
For both online and offline inference, the TAR module utilizes an attention window of 10 frames unless otherwise specified.
One-to-many Optimal Transport Assignment (OTA) is utilized instead of the traditional one-to-one Hungarian matcher for ground-truth matching for frame-level losses.

We use large-scale jittering (LSJ) for both VIS training and video pretraining. On VIS training, the same LSJ is applied to both frames preserving the consistency between objects across frames. On MS-COCO pseudo video pretraining, the same LSJ operation is applied independently over the two training frames, potentially leading to two very different-looking frames, which is desirable for pseudo videos. 

In video pretraining and VIS training phases, We train our model with resolutions up to 1024p for models with Resnet-50 backbones and resolutions up to 1280p for Swin-L backbones. When training models with Swin-L backbones, the number of feature levels of the backbone for the decoder is increased from 4 to 5, and the number of decoder transformer layers is increased from 6 to 9, which was shown to improve performance in MaskDINO.
The number of CDN queries is 100 for Resnet-50 models and 500 for Swin-L models.

MS-COCO image segmentation pretraining follows the settings of MaskDINO.
For the other two phases, the initial learning rate is set to $10 ^{-5}$ for the backbone and $10 ^{-4}$ for the rest of the network. 
For MS-COCO pseudo videos, the model is trained for 30k steps, with the learning rate being decreased to a tenth after 15k steps and again after 25k steps.
For the VIS training phase on all datasets, the model is trained for 13k iterations, with the learning rate being decreased to a tenth at the 10k iteration.

\subsection{Main results}

\begin{table*}[t]
	\small
	\centering
	\footnotesize
    \setlength{\tabcolsep}{5pt}
	\begin{tabular}{l|l|ccccc|ccccc|ccccc}
		\toprule
		\multicolumn{2}{l|}{\multirow{2}{*}{Method}} & \multicolumn{5}{c|}{YouTubeVIS-2019} & \multicolumn{5}{c|}{YouTubeVIS-2021} & \multicolumn{5}{c}{OVIS} \\
		\multicolumn{2}{l|}{} & AP & $\text{AP}_{50}$ & $\text{AP}_{75}$ & $\text{AR}_1$ & $\text{AR}_{10}$ & AP & $\text{AP}_{50}$ & $\text{AP}_{75}$ & $\text{AR}_1$ & $\text{AR}_{10}$ & AP & $\text{AP}_{50}$ & $\text{AP}_{75}$ & $\text{AR}_1$ & $\text{AR}_{10}$\\
		\midrule
		\parbox[t]{2mm}{\multirow{6}{*}{\rotatebox[origin=c]{90}{ResNet-50}}}
	    & SeqFormer~\cite{wu2021seqformer} & 47.4 & 69.8 & 51.8 & 45.5 & 54.8 & 40.5 & 62.4 & 43.7  & 36.1 & 48.1 & 15.1 & 31.9 & 13.8 & 10.4 & 27.1 \\
		& MinVIS~\cite{huang2022minvis} & 47.4 & 69.0 & 52.1 & 45.7 & 55.7 & 44.2 & 66.0 & 48.1 & 39.2 & 51.7 & 25.0 & 45.5 & 24.0 & 13.9 & 29.7 \\
		& VITA~\cite{heo2022vita} & 49.8 & 72.6 & 54.5 & \textbf{49.4} & \textbf{61.0} & 45.7 & 67.4 & 49.5 & 40.9 & 53.6 & 19.6 & 41.2 & 17.4 & 11.7 & 26.0 \\
		& IDOL~\cite{IDOL} & 49.5 & 74.0 & 52.9 & 47.7 & 58.7 & 43.9 & 68.0 & 49.6 & 38.0 & 50.9 & 30.2 & 51.3  & 30.0  &15.0 & 37.5 \\
		& \textbf{RefineVIS\textsubscript{online}} & 49.4 & 74.4 & 53.8 & 46.0 & 56.4 & 48.9 & 72.6 & 54.0 & \textbf{54.7} & 54.7 & 33.4 & \textbf{57.1} & 32.6 & \textbf{16.3} & \textbf{40.2} \\
		& \textbf{RefineVIS\textsubscript{offline}} & \textbf{52.2} & \textbf{76.3} & \textbf{57.7} & 47.5 & 57.6 & \textbf{50.2} & \textbf{72.8} & \textbf{55.4} & 41.2 & \textbf{56.3} & \textbf{33.7} & 56.2 & \textbf{34.8} & 15.6 & 39.8 \\
		
        \midrule
		\parbox[t]{2mm}{\multirow{6}{*}{\rotatebox[origin=c]{90}{Swin-L}}}
		& SeqFormer~\cite{wu2021seqformer} & 59.4 & 82.1 & 66.4 & 51.7 & 64.4 & 51.8 & 74.6 & 58.2 & 42.8 & 58.1  & 15.1 & 31.9 &13.8 & 10.4 & 27.1 \\
        & MinVIS~\cite{huang2022minvis} & 61.6 & 83.3 & 68.6 & 54.8 & 66.6 & 55.3 & 76.6 & 62.0 & 45.9 & 60.8 & 39.4 & 61.5 & 41.3 & 18.1 & 43.3 \\
		& VITA~\cite{heo2022vita} & 63.0  & 86.9  & 67.9 & \textbf{56.3}  & 68.1 & 57.5 & 80.6 & 61.0 & 47.7 & 62.6 & 27.7 & 51.9 & 24.9 & 14.9 & 33.0\\
        & IDOL~\cite{IDOL} & 64.3 &  87.5 & 71.0 & 55.6 & \textbf{69.1} & 56.1 & 80.8 & 63.5 & 45.0 & 60.1 & 42.6 & 65.7 &  45.2 & 17.9  & 49.6 \\
		& \textbf{RefineVIS\textsubscript{online}} & 64.3 & 87.6 & 70.9 & 55.8 & 68.2 & \textbf{61.4} & \textbf{84.1} & 68.5 & \textbf{48.3} & \textbf{65.2} & \textbf{46.1} & 69.7 & 47.8 & 19.0 & 50.8 \\
		& \textbf{RefineVIS\textsubscript{offline}} & \textbf{64.4} & \textbf{88.3} & \textbf{72.2} & 55.8 & 68.4 & 61.2 & 83.7 & \textbf{69.2} & 47.9 & 64.8 & 46.0 & \textbf{70.4} & \textbf{48.4} & \textbf{19.1} & \textbf{51.2} \\
		\bottomrule
	\end{tabular}
	\caption{Evaluation results on \textbf{YouTubeVIS-2019}, \textbf{YouTubeVIS-2021} and \textbf{OVIS} validation splits.}
	\label{tab:ytvis2021}
\end{table*}

For inference, the smaller size of the input frames is resized to 480p if it is a ResNet-50 model, 720p if Swin-L.
In Table \ref{sec:experiments}, we compare our model, both in online and offline mode, with other recent works on the three datasets.
Unless otherwise noted, all online and offline inference results used a refinement window of 10 frames, and all models were first pretrained with both regular MS-COCO and pseudo videos.

Across all the datasets, models with ResNet-50 backbone benefit from offline inference compared to online inference, while models with Swin-L backbone have a similar result.
The reason is that the frame-level predictions are noisier when a weaker backbone is utilized. The offline inference can use the information from the frames ahead of the current frame to reduce some noise.

\textbf{YouTubeVIS-2019.} This is the easiest VIS dataset because it only contains short videos. The online variant of our model performs similarly to IDOL, because most of the masks in this dataset can be predicted accurately using frame-level models. However, our offline variant performs better than both IDOL and VITA on these datasets.

\textbf{YoutubeVIS-2021.} On this more extensive dataset, we see a more pronounced advantage of our RefineVIS compared to other models. The online variant delivers 3.2 and 3.9 and extra AP points than the previous best method with ResNet-50 and Swin-L backbones, respectively. Our offline variant brings 4.5 and 3.9 extra AP points than VITA, the previous best method, with ResNet-50 and Swin-L backbones, respectively.

\textbf{OVIS} is the most challenging dataset in the benchmark. This dataset also shows a clear advantage of the proposed framework.
When compared with IDOL, the previous leader, our scores are better by 3.5 AP points with both ResNet-50 and Swin-L.
We also notice that our online and offline methods perform similarly on this dataset.


\subsection{Ablation studies}

We focus our ablation studies on the OVIS and on the YouTube-2021 datasets, providing experimental results that consistently show the importance of each element that we introduce and how they provide state-of-the-art VIS results.

\textbf{Video Contrastive Denoising}. Table \ref{tab:ablation_denoising} shows the incremental gains of using the new CDN applications over OVIS models that were not pretrained with MS-COCO pseudo videos.
In the first row, we see the results of only using CDN at the frame level, like the standard MaskDINO.
We can see a gain of $0.9$ AP after applying temporal refinement.
In the second row, we see the gains from just introducing Association CDN, which benefits both frame-level and refinement parts of our model and improves the frame-level AP by $1.1$ and the refined AP by $0.8$.
Again we see a gain in using refinement: $0.6$ AP.
Finally, we see a gain of $0.5$ AP by introducing Temporal CDN and improving refined mask predictions.
Temporal CDN does not affect frame-level results.

\begin{table}
    \centering
    \footnotesize
    \setlength{\tabcolsep}{5pt}
    \begin{tabular}{l|ccc|ccc}
        \toprule
        \multirow{2}{*}{Training mode} & \multicolumn{3}{c|}{Frame-level results} & \multicolumn{3}{c}{Refined results} \\
        & AP & $\text{AP}_{50}$ & $\text{AP}_{75}$ & AP & $\text{AP}_{50}$ & $\text{AP}_{75}$ \\
		\midrule
        Frame CDN         & 30.1 & 51.3 & 29.6 & 31.0 & 53.1 & 29.3 \\
        + Association CDN & 31.2 & 52.0 & 30.7 & 31.8 & 54.2 & 30.8 \\
        + Temporal CDN    & -    & -    & -    & 32.3 & 55.5 & 31.7 \\
    \bottomrule
    \end{tabular}
    \caption{Cumulative contributions of the CDN technique to both frame-level and video-level parts of the model. Offline results on the OVIS dataset of ResNet-50 models without MS-COCO pseudo video pretraining. Frame CDN is the original used by MaskDINO.}
    \label{tab:ablation_denoising}
\end{table}

\textbf{Temporal Attention Window}. We evaluate the choice of $W$, the refinement window length, in Table \ref{tab:ablation_attn_win}. 
In this experiment, we experiment with different refinement window sizes from 2 to 20. A larger window size leads to better AP until a window size of 10. Using windows that are too small or too large decreases the AP.
Small windows add noise to the predictions, while large ones will cause object appearance to change too much inside the refinement window.
  
\begin{table}
    \centering 
    \footnotesize 
    \begin{tabular}{c|c|cccccc} 
        \toprule 
        $W$ & 0 & 2 & 4 & 8 & 10 & 16 & 20 \\ 
        \midrule 
        AP & 32.7 & 32.1 & 32.9 & 33.5 & 33.7 & 33.3 & 32.9 \\ 
    \bottomrule 
    \end{tabular} 
    \caption{OVIS offline results with a ResNet-50 showing how different values of $W$ (the length of the window consumed by TAR) impact the overall performance. $0$ means refinement is disabled.} 
    \label{tab:ablation_attn_win} 
\end{table}

\textbf{Training with more frames.} Our study into the effects of training RefineVIS with additional frames is presented in Table \ref{tab:nframes_train}.
Unfortunately, using more frames leads to a decrease in AP, which our investigation suggests comes from instability in the training loss.
Another challenge from using more frames during training is the considerable increase in memory usage by the CDN technique, which requires accommodating hundreds of extra transformer queries.
Addressing these challenges remains an open research topic.

\begin{table}
    \centering
    \footnotesize
    \setlength{\tabcolsep}{5pt}
    \begin{tabular}{c|l|ccccc}
        \toprule
        \# of frames & AP & $\text{AP}_{50}$ & $\text{AP}_{75}$ & $\text{AR}_{1}$ & $\text{AR}_{10}$ \\
        \midrule
        2 & 50.2 & 72.8 & 55.4 & 41.2 & 56.3 \\
        3 & 49.9 & 72.8 & 55.5 & 41.2 & 55.3 \\
        5 & 49.4 & 72.1 & 55.0 & 41.1 & 55.5 \\
    \bottomrule
    \end{tabular}
    \caption{Effect of using different number of frames during training. Results from a ResNet-50 backbone on YouTubeVIS-2021.}
    \label{tab:nframes_train}
\end{table}

\textbf{Temporal Refinement Module}. Tables \ref{tab:ablation_denoising} and \ref{tab:ablation_attn_win} show temporal refinement module consistently delivers better results than the baseline frame-level model under different training schemes.
Table \ref{tab:ablation_ovis_swinl} shows another ablation study that highlights the impact of TAR on YouTubeVIS-2021.
Results are from RefineVIS with a ResNet-50 backbone both with and without TAR.
The same table also measures the impact of Large Scale Jittering (LSJ) to the training process and how TAR and LSJ synergistically contribute to the performance of RefineVIS.

\begin{table}
    \centering
    \footnotesize
    \begin{tabular}{cc|ccccc}
        \toprule
        TAR & LSJ & AP & $\text{AP}_{50}$ & $\text{AP}_{75}$ & $\text{AR}_{1}$ & $\text{AR}_{10}$ \\
        \midrule
        &                       & 48.1 & 70.5 & 53.0 & 39.3 & 53.2 \\
        & \checkmark            & 48.9 & 73.0 & 53.1 & 40.7 & 54.9 \\
        \checkmark &            & 49.1 & 71.9 & 54.1 & 39.8 & 54.3 \\
        \checkmark & \checkmark & 50.2 & 72.8 & 55.4 & 41.2 & 56.3 \\
        
    \bottomrule
    \end{tabular}
    \caption{YouTubeVIS-2021 offline results with a ResNet-50 backbone highlight the impact of TAR and LSJ.}
    \label{tab:ablation_ovis_swinl}
\end{table}

\textbf{Inference speed.} Table \ref{tab:ablation_fps} compares the speed of our ResNet-50 model, with and without TAR.
Numbers are the average FPS for the models to process the entire YouTubeVIS-2021 dataset on a NVIDIA V-100 GPU.
Regardless of the backbone, TAR adds only 2.8 MFLOPS per 10-frame window, which represents a minor impact.

\begin{table}
    \centering
    \footnotesize
    \setlength{\tabcolsep}{5pt}
    \begin{tabular}{c|cc}
        \toprule
        TAR & ResNet-50 &  Swin-L \\
        \midrule
                   & 22.27 & 4.63 \\
        \checkmark & 21.91 & 4.60 \\
    \bottomrule
    \end{tabular}
    \caption{FPS of different configurations on a V-100 GPU}
    \label{tab:ablation_fps}
\end{table}

\textbf{RefineVIS as a framework.} To further demonstrate the modularity of RefineVIS, we conduct studies with two other segmentation models.
The first uses Mask2Former \cite{cheng2021mask2former} and the second a CondInst \cite{tian2020conditional} mask head with Deformable DETR transformer \cite{zhu2020deformable}, similar to the one in IDOL.
We begin with pre-trained versions of these segmentation models and finetune on YouTubeVIS-2021 under the RefineVIS framework.
Results are shown in Table \ref{tab:framework} and demonstrate how much our framework can enhance these spatial-only segmentation models.

\begin{table}
    \centering
    \footnotesize
    \setlength{\tabcolsep}{5pt}
    \begin{tabular}{l|c|cccccc}
        \toprule
        Segm. model & TAR & AP & $\text{AP}_{50}$ & $\text{AP}_{75}$ & $\text{AR}_{1}$ & $\text{AR}_{10}$ \\
        \midrule
        \multirow{2}{*}{Mask2Former} & & 44.0 & 67.4 & 47.4 & 38.9 & 51.5 \\
                          & \checkmark & 45.2 & 69.3 & 49.8 & 39.5 & 52.6 \\
        \midrule
        \multirow{2}{*}{CondInst} & & 47.8 & 71.4 & 52.9 & 41.2 & 55.6 \\
                       & \checkmark & 48.3 & 72.4 & 53.9 & 41.4 & 55.3 \\
    \end{tabular}
    \caption{Results of other segmentation models trained under the RefineVIS framework show TAR is effective and the framework is extensible.}
    \label{tab:framework}
\end{table}

\section{Conclusion}
\label{sec:conclusion}

In this work, we propose a video instance segmentation framework called RefineVIS.
It can associate objects in long videos and employ temporal information to refine mask predictions, leading to more accurate predictions in challenging cases.
RefineVIS supports both online and offline inference modes. It achieves state-of-the-art in both modes on YouTubeVIS-2019, YouTubeVIS-2021, and OVIS datasets.

{\small
\bibliographystyle{ieee_fullname}
\bibliography{11_references}

\begin{thebibliography}{10}\itemsep=-1pt

\bibitem{cao2020sipmask}
Jiale Cao, Rao~Muhammad Anwer, Hisham Cholakkal, Fahad~Shahbaz Khan, Yanwei
  Pang, and Ling Shao.
\newblock Sipmask: Spatial information preservation for fast image and video
  instance segmentation.
\newblock In {\em European Conference on Computer Vision}, pages 1--18.
  Springer, 2020.

\bibitem{cheng2021mask2formervideo}
Bowen Cheng, Anwesa Choudhuri, Ishan Misra, Alexander Kirillov, Rohit Girdhar,
  and Alexander~G Schwing.
\newblock Mask2former for video instance segmentation.
\newblock {\em arXiv preprint arXiv:2112.10764}, 2021.

\bibitem{cheng2021mask2former}
Bowen Cheng, Ishan Misra, Alexander~G. Schwing, Alexander Kirillov, and Rohit
  Girdhar.
\newblock Masked-attention mask transformer for universal image segmentation.
\newblock 2022.

\bibitem{han2022visolo}
Su~Ho Han, Sukjun Hwang, Seoung~Wug Oh, Yeonchool Park, Hyunwoo Kim, Min-Jung
  Kim, and Seon~Joo Kim.
\newblock Visolo: Grid-based space-time aggregation for efficient online video
  instance segmentation.
\newblock In {\em Proceedings of the IEEE/CVF Conference on Computer Vision and
  Pattern Recognition}, pages 2896--2905, 2022.

\bibitem{inspro}
Fei He, Haoyang Zhang, Naiyu Gao, Jian Jia, Yanhu Shan, Xin Zhao, and Kaiqi
  Huang.
\newblock Inspro: Propagating instance query and proposal for online video
  instance segmentation.
\newblock In S. Koyejo, S. Mohamed, A. Agarwal, D. Belgrave, K. Cho, and A. Oh,
  editors, {\em Advances in Neural Information Processing Systems}, volume~35,
  pages 19370--19383. Curran Associates, Inc., 2022.

\bibitem{he2017mask}
Kaiming He, Georgia Gkioxari, Piotr Doll{\'a}r, and Ross Girshick.
\newblock Mask r-cnn.
\newblock In {\em Proceedings of the IEEE international conference on computer
  vision}, pages 2961--2969, 2017.

\bibitem{heo2022vita}
Miran Heo, Sukjun Hwang, Seoung~Wug Oh, Joon-Young Lee, and Seon~Joo Kim.
\newblock Vita: Video instance segmentation via object token association.
\newblock 2022.

\bibitem{huang2022minvis}
De-An Huang, Zhiding Yu, and Anima Anandkumar.
\newblock Minvis: A minimal video instance segmentation framework without
  video-based training.
\newblock 2022.

\bibitem{hwang2021video}
Sukjun Hwang, Miran Heo, Seoung~Wug Oh, and Seon~Joo Kim.
\newblock Video instance segmentation using inter-frame communication
  transformers.
\newblock {\em Advances in Neural Information Processing Systems}, 34, 2021.

\bibitem{ke2021prototypical}
Lei Ke, Xia Li, Martin Danelljan, Yu-Wing Tai, Chi-Keung Tang, and Fisher Yu.
\newblock Prototypical cross-attention networks for multiple object tracking
  and segmentation.
\newblock {\em Advances in Neural Information Processing Systems},
  34:1192--1203, 2021.

\bibitem{li2022maskdino}
Feng Li, Hao Zhang, Huaizhe xu, Shilong Liu, Lei Zhang, Lionel~M. Ni, and
  Heung-Yeung Shum.
\newblock Mask dino: Towards a unified transformer-based framework for object
  detection and segmentation, 2022.

\bibitem{li2021spatial}
Minghan Li, Shuai Li, Lida Li, and Lei Zhang.
\newblock Spatial feature calibration and temporal fusion for effective
  one-stage video instance segmentation.
\newblock In {\em Proceedings of the IEEE/CVF Conference on Computer Vision and
  Pattern Recognition}, pages 11215--11224, 2021.

\bibitem{mscoco}
Tsung{-}Yi Lin, Michael Maire, Serge~J. Belongie, Lubomir~D. Bourdev, Ross~B.
  Girshick, James Hays, Pietro Perona, Deva Ramanan, Piotr Doll{\'{a}}r, and
  C.~Lawrence Zitnick.
\newblock Microsoft {COCO:} common objects in context.
\newblock {\em CoRR}, abs/1405.0312, 2014.

\bibitem{liu2021sg}
Dongfang Liu, Yiming Cui, Wenbo Tan, and Yingjie Chen.
\newblock Sg-net: Spatial granularity network for one-stage video instance
  segmentation.
\newblock {\em Proceedings of the IEEE/CVF Conference on Computer Vision and
  Pattern Recognition}, 2021.

\bibitem{qi2022occluded}
Jiyang Qi, Yan Gao, Yao Hu, Xinggang Wang, Xiaoyu Liu, Xiang Bai, Serge
  Belongie, Alan Yuille, Philip~HS Torr, and Song Bai.
\newblock Occluded video instance segmentation: A benchmark.
\newblock {\em International Journal of Computer Vision}, 130(8), 2022.

\bibitem{mssts}
Omkar Thawakar, Sanath Narayan, Jiale Cao, Hisham Cholakkal, Rao~Muhammad
  Anwer, Muhammad~Haris Khan, Salman Khan, Michael Felsberg, and Fahad~Shahbaz
  Khan.
\newblock Video instance segmentation via multi-scale spatio-temporal split
  attention transformer.
\newblock In {\em Computer Vision – ECCV 2022: 17th European Conference, Tel
  Aviv, Israel, October 23–27, 2022, Proceedings, Part XXIX}, page 666–681,
  Berlin, Heidelberg, 2022. Springer-Verlag.

\bibitem{tian2020conditional}
Zhi Tian, Chunhua Shen, and Hao Chen.
\newblock Conditional convolutions for instance segmentation.
\newblock In {\em European Conference on Computer Vision}, pages 282--298.
  Springer, 2020.

\bibitem{vaswani2017attention}
Ashish Vaswani, Noam Shazeer, Niki Parmar, Jakob Uszkoreit, Llion Jones,
  Aidan~N Gomez, {\L}ukasz Kaiser, and Illia Polosukhin.
\newblock Attention is all you need.
\newblock {\em Advances in neural information processing systems}, 30, 2017.

\bibitem{wang2021end}
Yuqing Wang, Zhaoliang Xu, Xinlong Wang, Chunhua Shen, Baoshan Cheng, Hao Shen,
  and Huaxia Xia.
\newblock End-to-end video instance segmentation with transformers.
\newblock In {\em Proceedings of the IEEE/CVF Conference on Computer Vision and
  Pattern Recognition}, pages 8741--8750, 2021.

\bibitem{wu2021seqformer}
Junfeng Wu, Yi Jiang, Song Bai, Wenqing Zhang, and Xiang Bai.
\newblock Seqformer: Sequential transformer for video instance segmentation.
\newblock In {\em ECCV}, 2022.

\bibitem{IDOL}
Junfeng Wu, Qihao Liu, Yi Jiang, Song Bai, Alan Yuille, and Xiang Bai.
\newblock In defense of online models for video instance segmentation.
\newblock In {\em ECCV}, 2022.

\bibitem{youtubevis2021}
Ning Xu, Linjie Yang, Yuchen Fan, Yang Fu, Lin Weiyao, Jianchao Yang, Humphrey
  Shi, Joon-Young Lee, and Seonguk Seo.
\newblock \url{https://youtube-vos.org/challenge/2021/}, 2021.
\newblock The 3rd Large-scale Video Object Segmentation Challenge.

\bibitem{yang2019video}
Linjie Yang, Yuchen Fan, and Ning Xu.
\newblock Video instance segmentation.
\newblock In {\em Proceedings of the IEEE/CVF International Conference on
  Computer Vision}, pages 5188--5197, 2019.

\bibitem{yang2021crossover}
Shusheng Yang, Yuxin Fang, Xinggang Wang, Yu Li, Chen Fang, Ying Shan, Bin
  Feng, and Wenyu Liu.
\newblock Crossover learning for fast online video instance segmentation.
\newblock In {\em Proceedings of the IEEE/CVF International Conference on
  Computer Vision}, pages 8043--8052, 2021.

\bibitem{yang2022temporally}
Shusheng Yang, Xinggang Wang, Yu Li, Yuxin Fang, Jiemin Fang, Wenyu Liu, Xun
  Zhao, and Ying Shan.
\newblock Temporally efficient vision transformer for video instance
  segmentation.
\newblock In {\em Proceedings of the IEEE/CVF Conference on Computer Vision and
  Pattern Recognition}, pages 2885--2895, 2022.

\bibitem{ifr2023}
Quanzeng You, Jiang Wang, Peng Chu, Andre Abrantes, and Zicheng Liu.
\newblock Consistent video instance segmentation with inter-frame recurrent
  attention, 2022.

\bibitem{zhang2022dino}
Hao Zhang, Feng Li, Shilong Liu, Lei Zhang, Hang Su, Jun Zhu, Lionel~M. Ni, and
  Heung-Yeung Shum.
\newblock Dino: Detr with improved denoising anchor boxes for end-to-end object
  detection, 2022.

\bibitem{taformer}
Zhenghao Zhang, Fangtao Shao, Zuozhuo Dai, and Siyu Zhu.
\newblock Towards robust video instance segmentation with temporal-aware
  transformer, 2023.

\bibitem{zhu2020deformable}
Xizhou Zhu, Weijie Su, Lewei Lu, Bin Li, Xiaogang Wang, and Jifeng Dai.
\newblock Deformable detr: Deformable transformers for end-to-end object
  detection.
\newblock In {\em International Conference on Learning Representations}, 2020.

\end{thebibliography}
}

\ifarxiv \clearpage \appendix
\label{sec:appendix}

\section{Qualitative evaluation}

The visualizations of OVIS videos presented in Figure \ref{fig:vis} demonstrate the real-world applicability and effectiveness of our RefineVIS framework, specially the TAR module.
They illustrate how our method utilizes temporal context to refine predictions in instance segmentation.
In the first two video examples, instances that were not detected initially were accurately captured after applying the TAR module.
The third video exemplifies how temporal refinement contributes to continuity in instance tracking, which is crucial for VIS.
Notably, the application of TAR ensured the consistency of instance masks over successive frames, providing more better  segmentation results.
These outcomes underscore the robustness of RefineVIS in handling challenging VIS scenarios and improving overall segmentation accuracy.

\begin{figure*}[t]
	\small
	
	\centering
    \renewcommand{\arraystretch}{0.5} 
        
        \begin{tabular}
		{@{\hspace{0.5mm}}c@{\hspace{0.5mm}}c@{\hspace{0.5mm}}c@{\hspace{.5mm}}c@{\hspace{.5mm}}c@{\hspace{.5mm}}}%
		\includegraphics[width=0.197\linewidth]{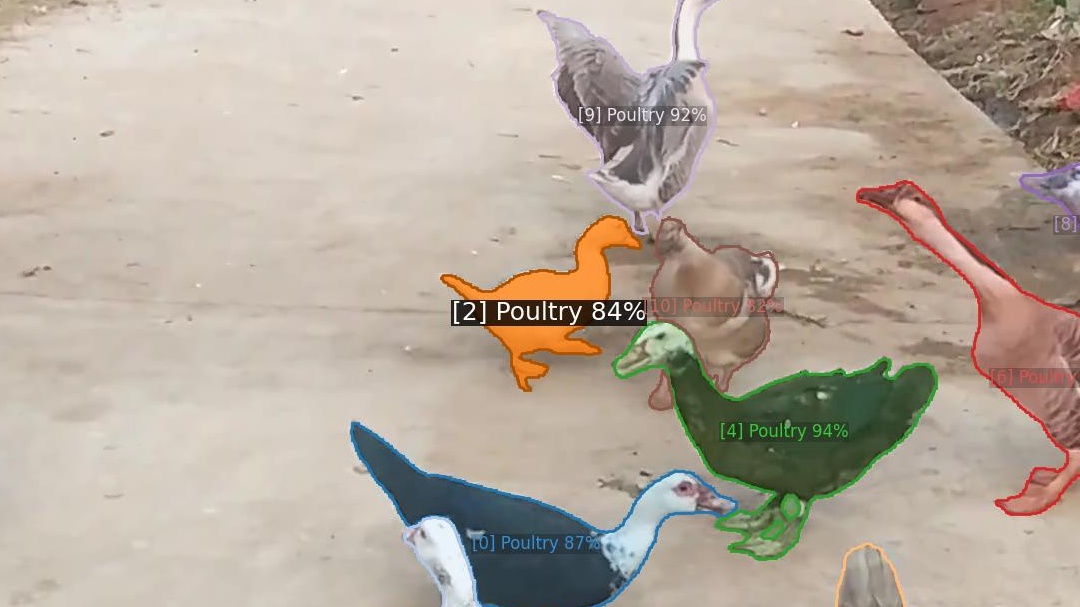}    & 
		\includegraphics[width=0.197\linewidth]{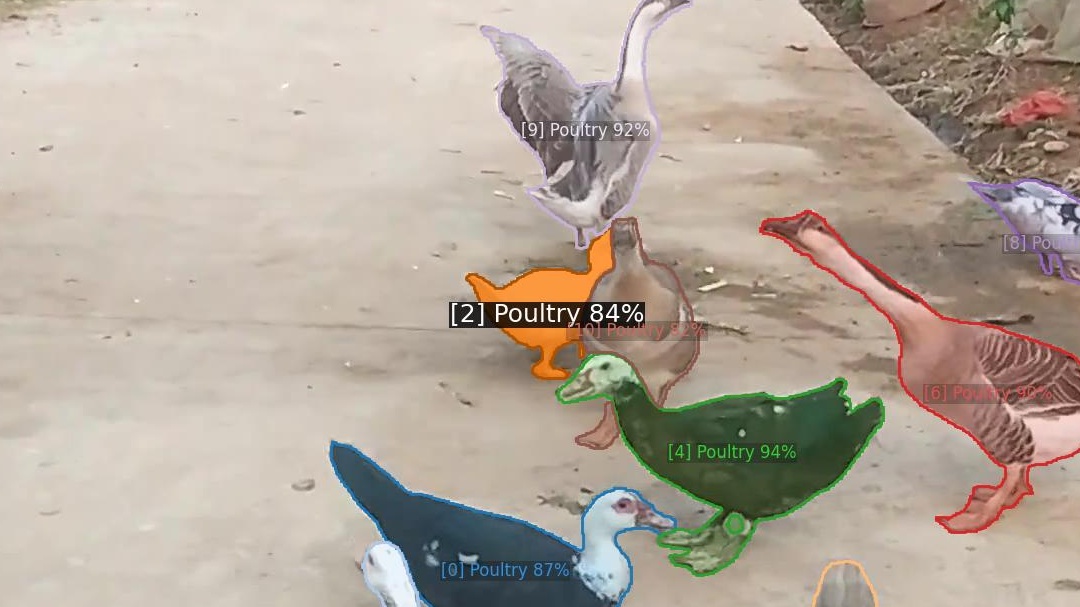}    & 
		\includegraphics[width=0.197\linewidth]{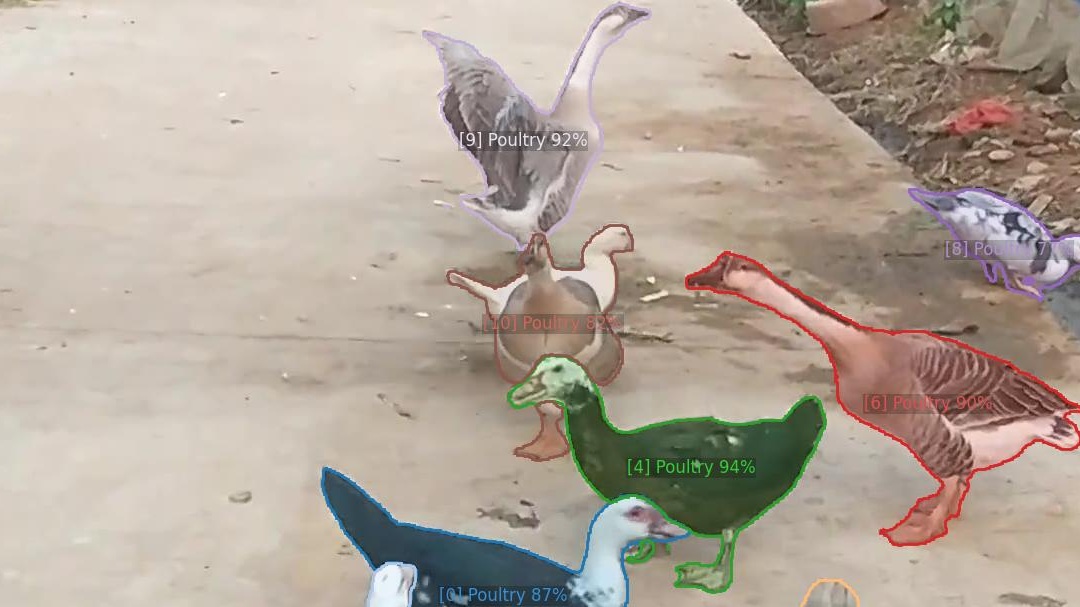}    &
		\includegraphics[width=0.197\linewidth]{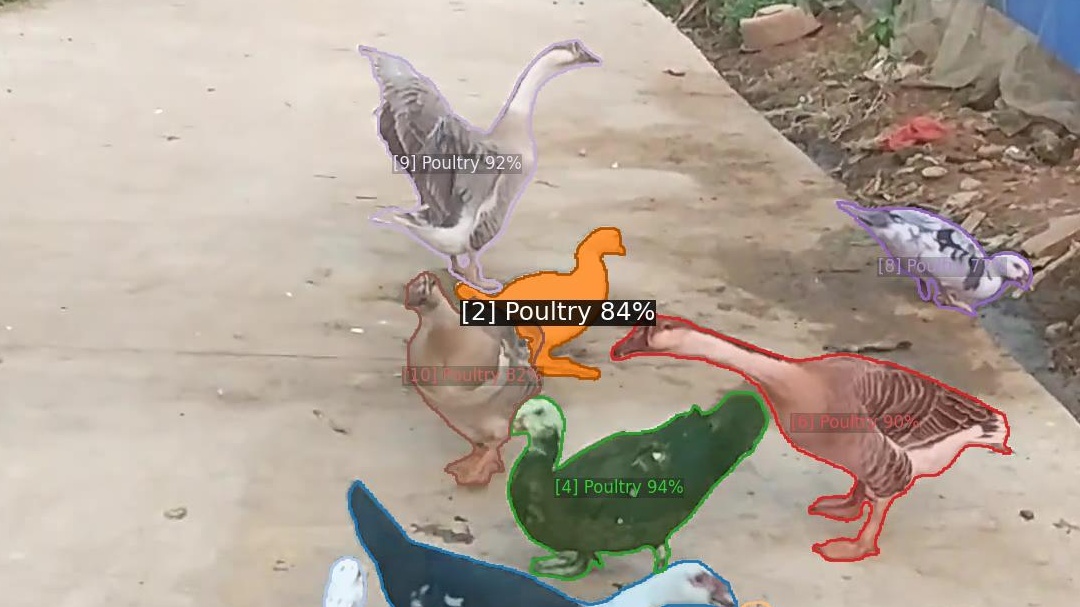} &
        \includegraphics[width=0.197\linewidth]{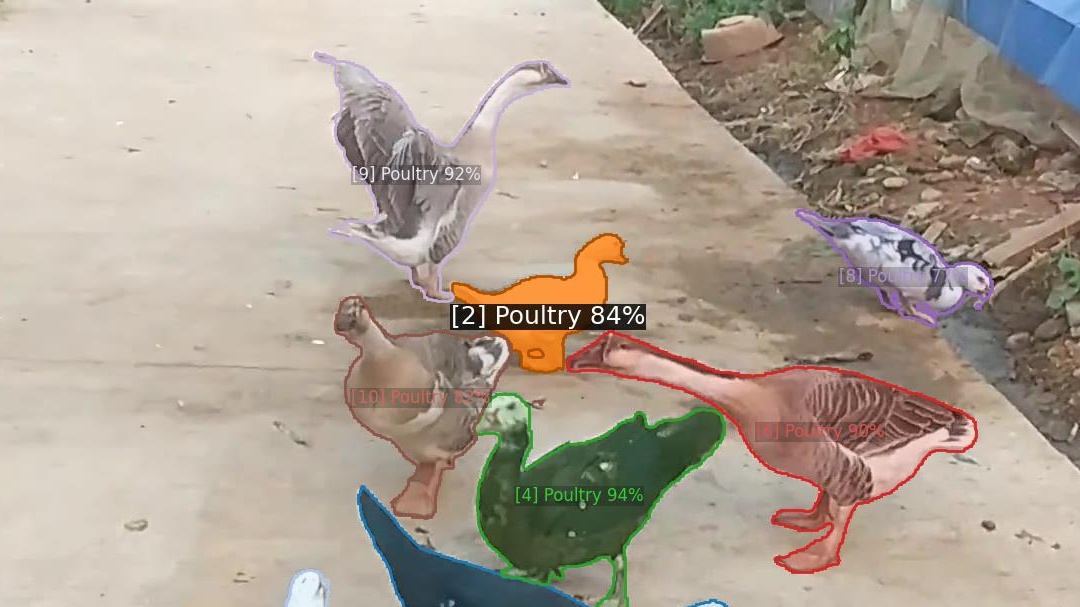}
		\\
		\includegraphics[width=0.197\linewidth]{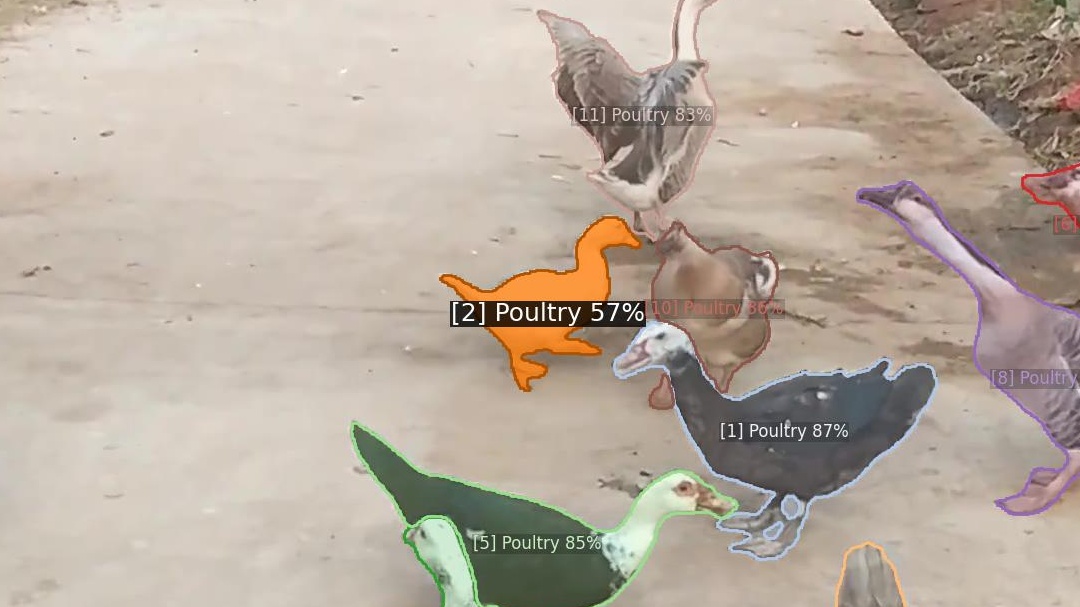}    & 
		\includegraphics[width=0.197\linewidth]{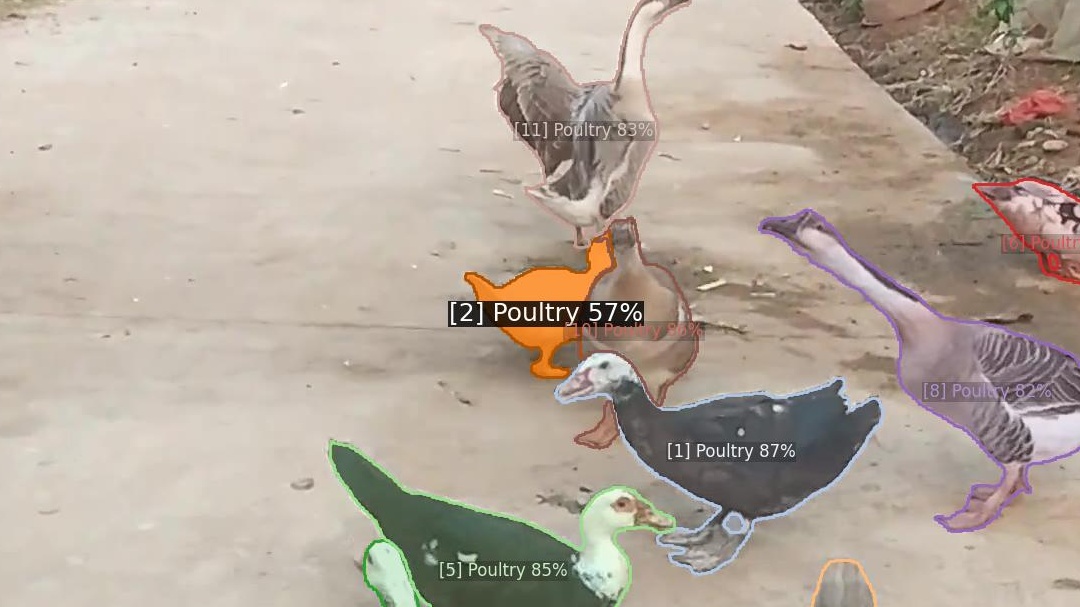}    & 
		\includegraphics[width=0.197\linewidth]{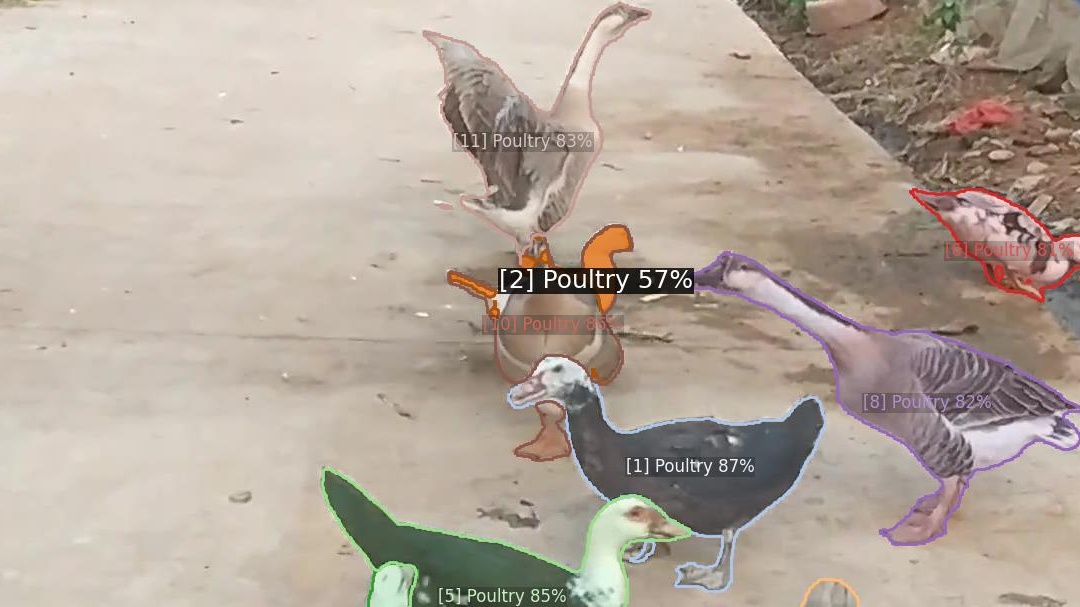}    &
		\includegraphics[width=0.197\linewidth]{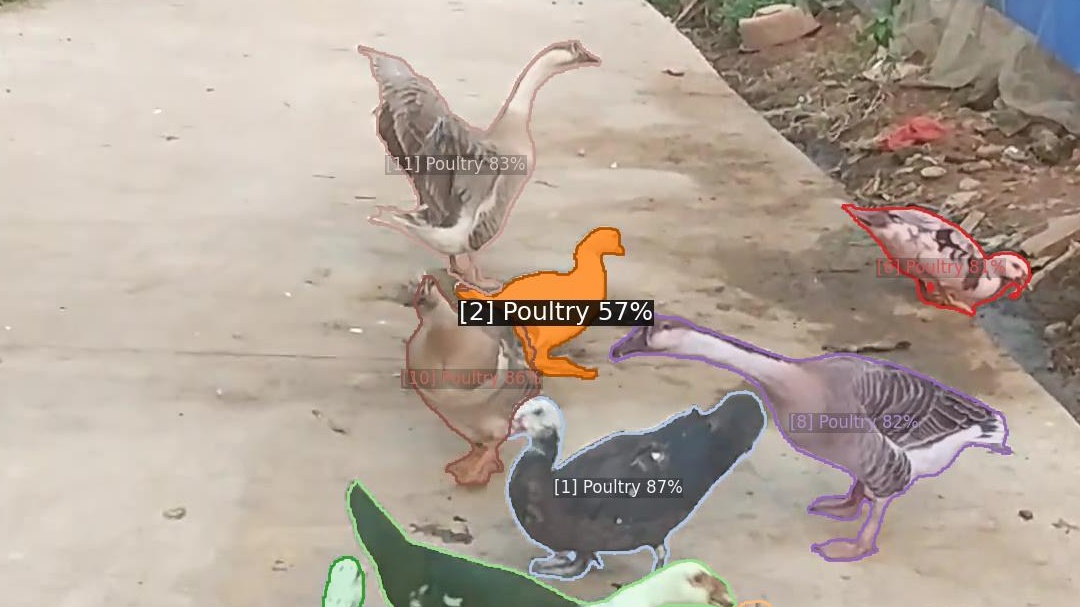} &
        \includegraphics[width=0.197\linewidth]{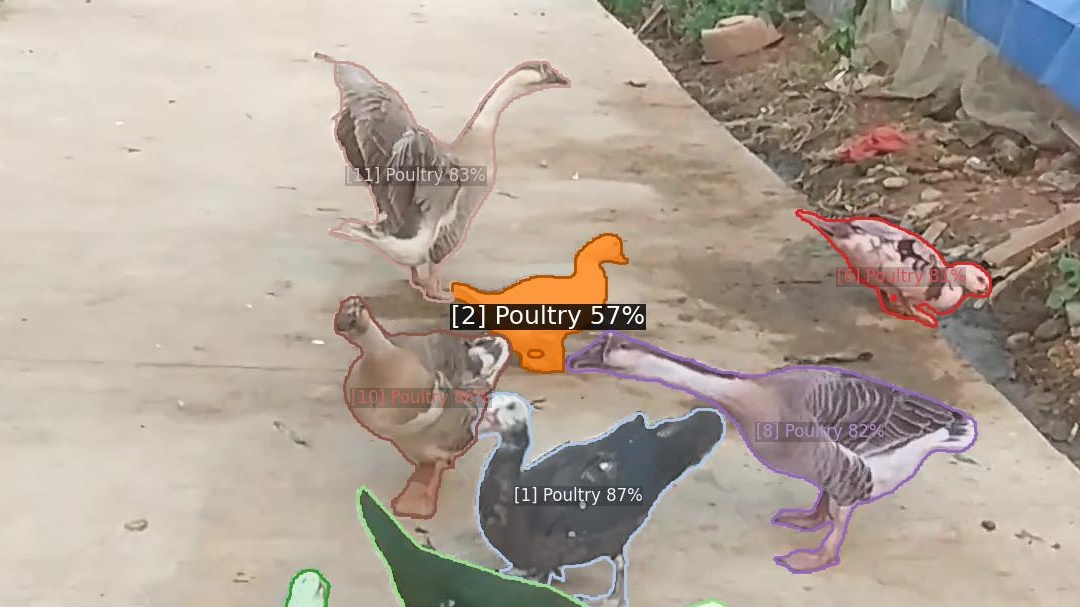}
		\end{tabular}

        \begin{tabular}
		{@{\hspace{0.5mm}}c@{\hspace{0.5mm}}c@{\hspace{0.5mm}}c@{\hspace{.5mm}}c@{\hspace{.5mm}}c@{\hspace{.5mm}}}%
		\includegraphics[width=0.197\linewidth]{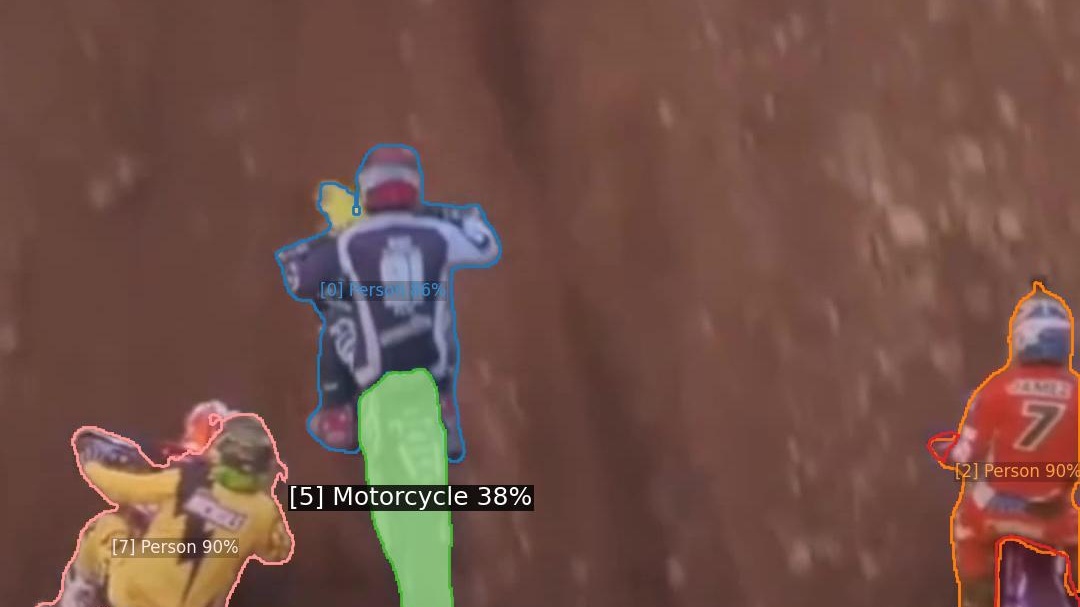}    & 
		\includegraphics[width=0.197\linewidth]{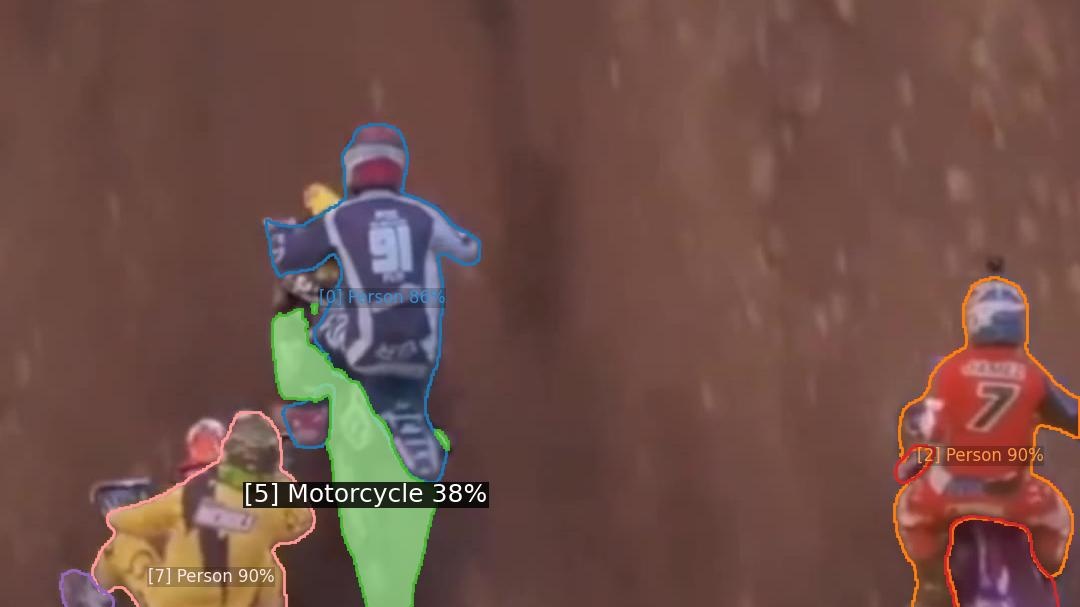}    & 
		\includegraphics[width=0.197\linewidth]{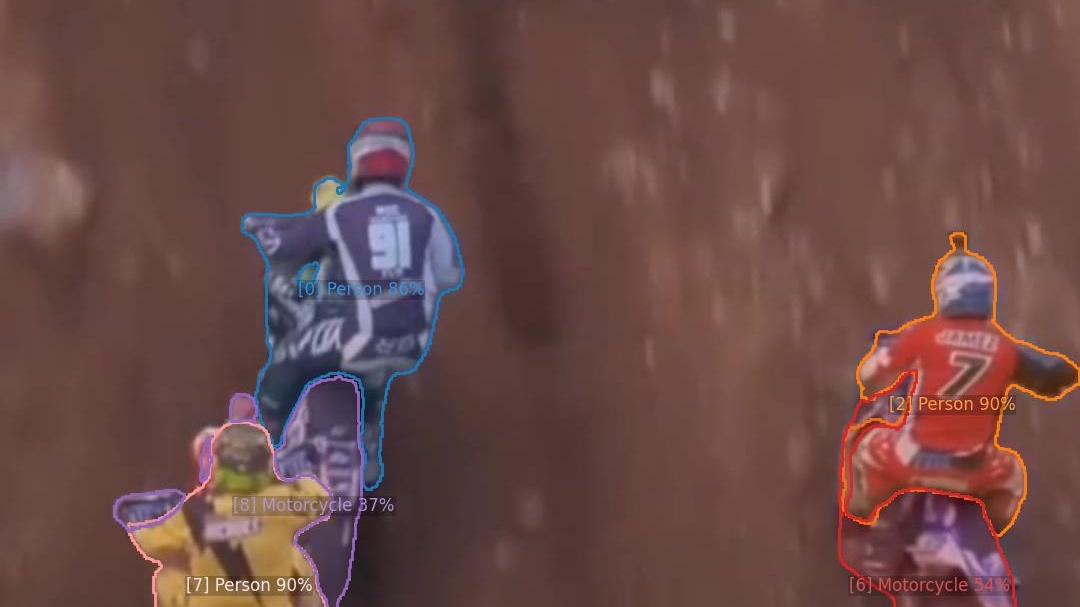}    &
		\includegraphics[width=0.197\linewidth]{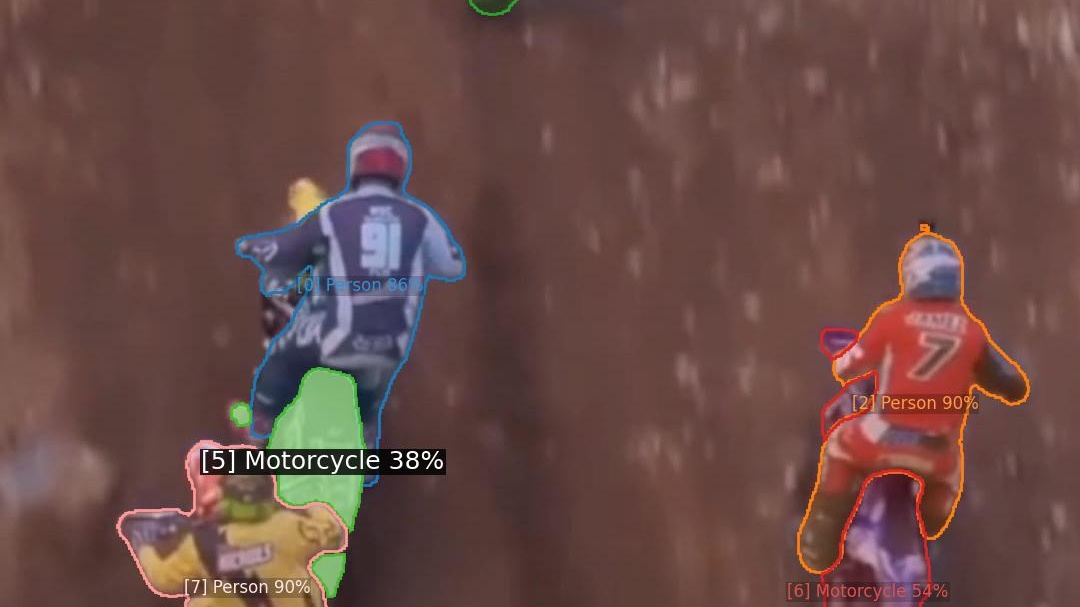} &
        \includegraphics[width=0.197\linewidth]{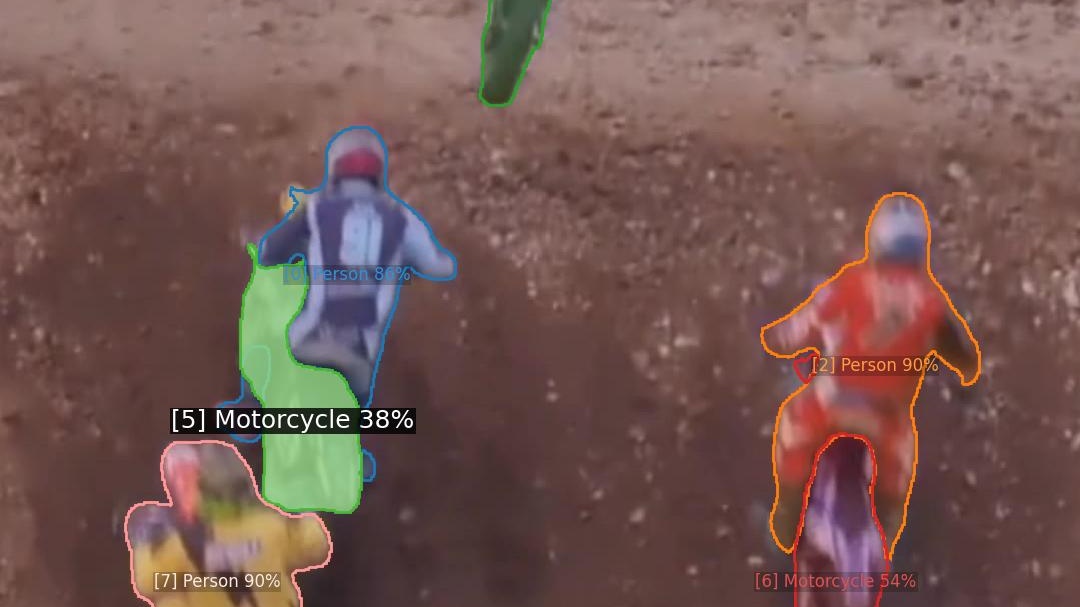}
		\\
		\includegraphics[width=0.197\linewidth]{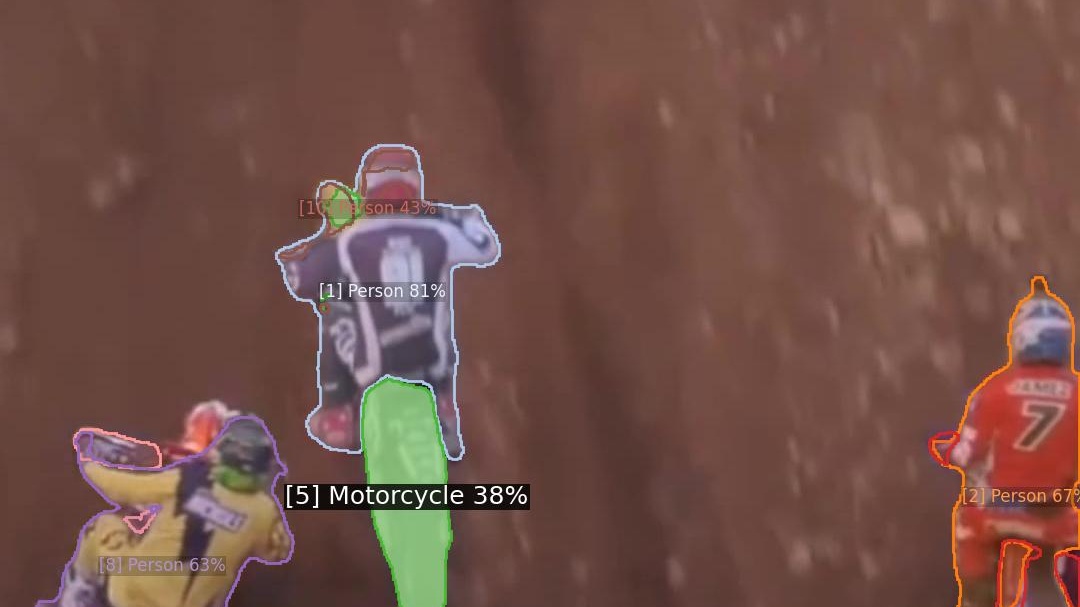}    & 
		\includegraphics[width=0.197\linewidth]{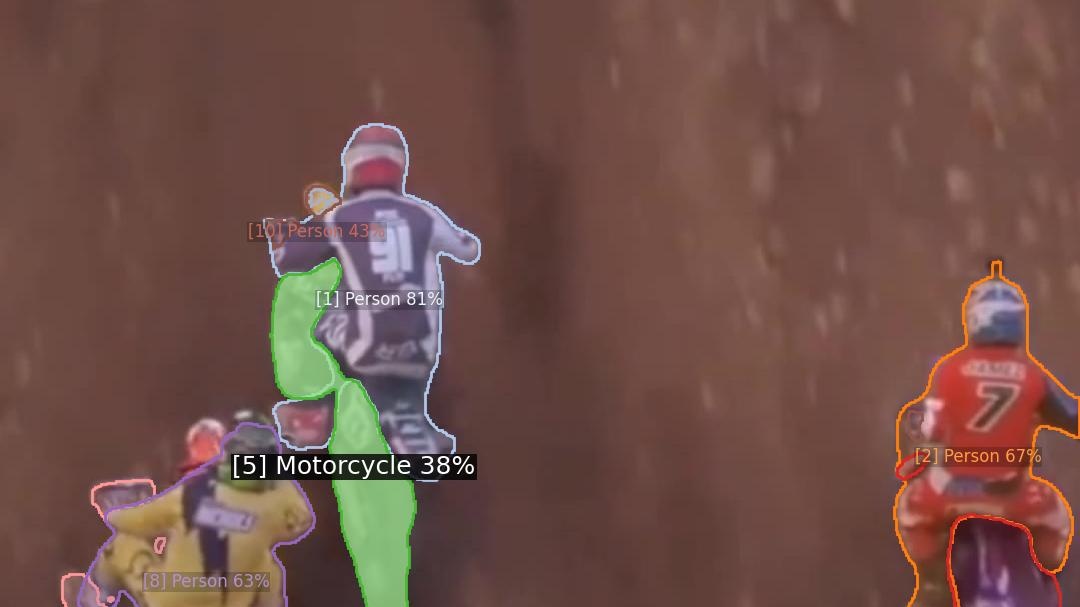}    & 
		\includegraphics[width=0.197\linewidth]{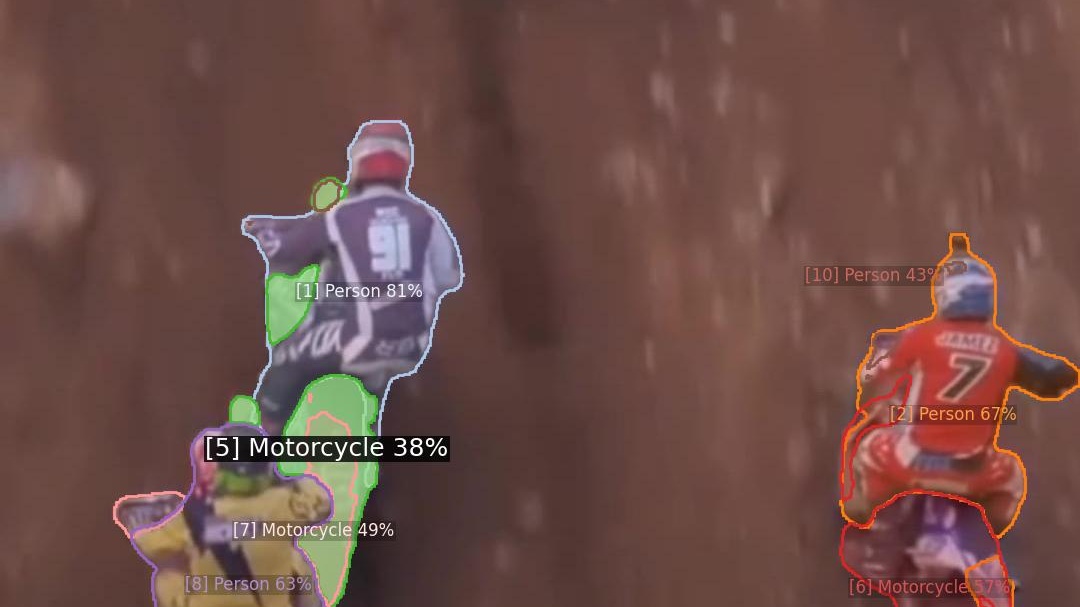}    &
		\includegraphics[width=0.197\linewidth]{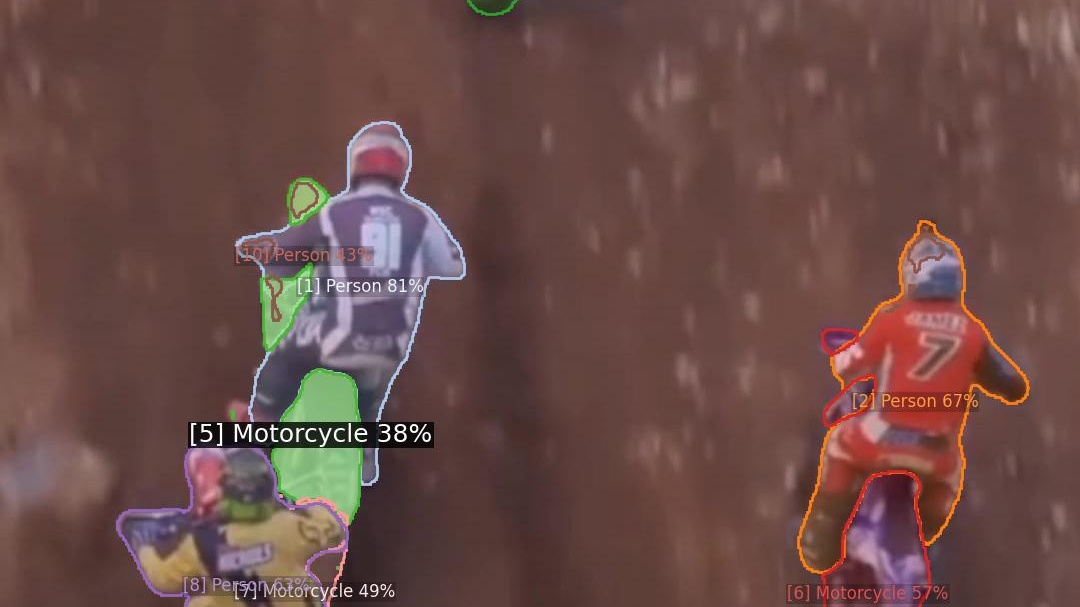} &
        \includegraphics[width=0.197\linewidth]{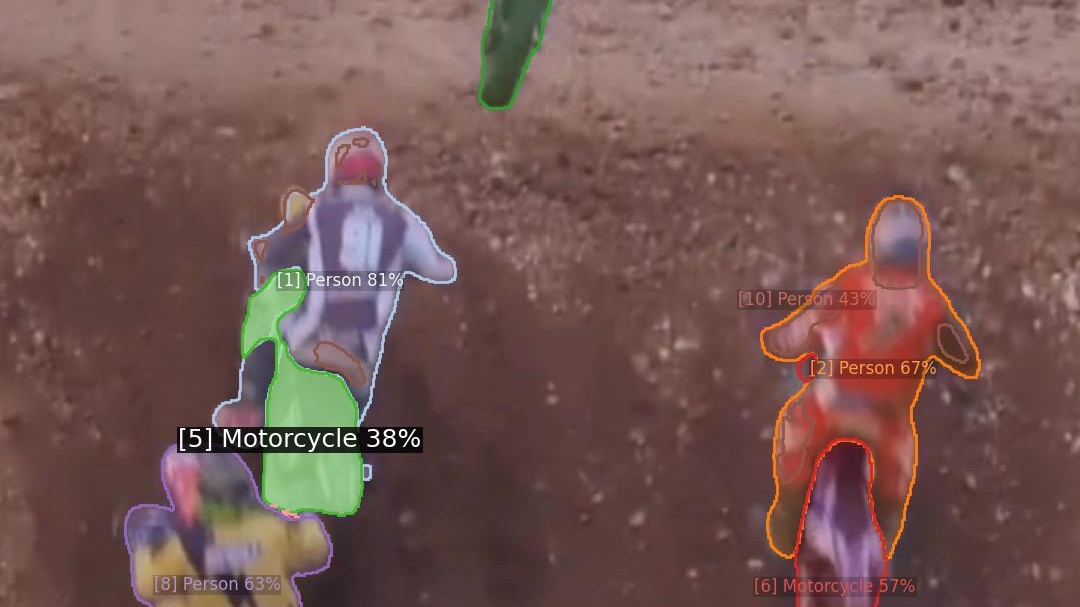}
		\end{tabular}
  
        \begin{tabular}
		{@{\hspace{0.5mm}}c@{\hspace{0.5mm}}c@{\hspace{0.5mm}}c@{\hspace{.5mm}}c@{\hspace{.5mm}}c@{\hspace{.5mm}}}%
		\includegraphics[width=0.197\linewidth]{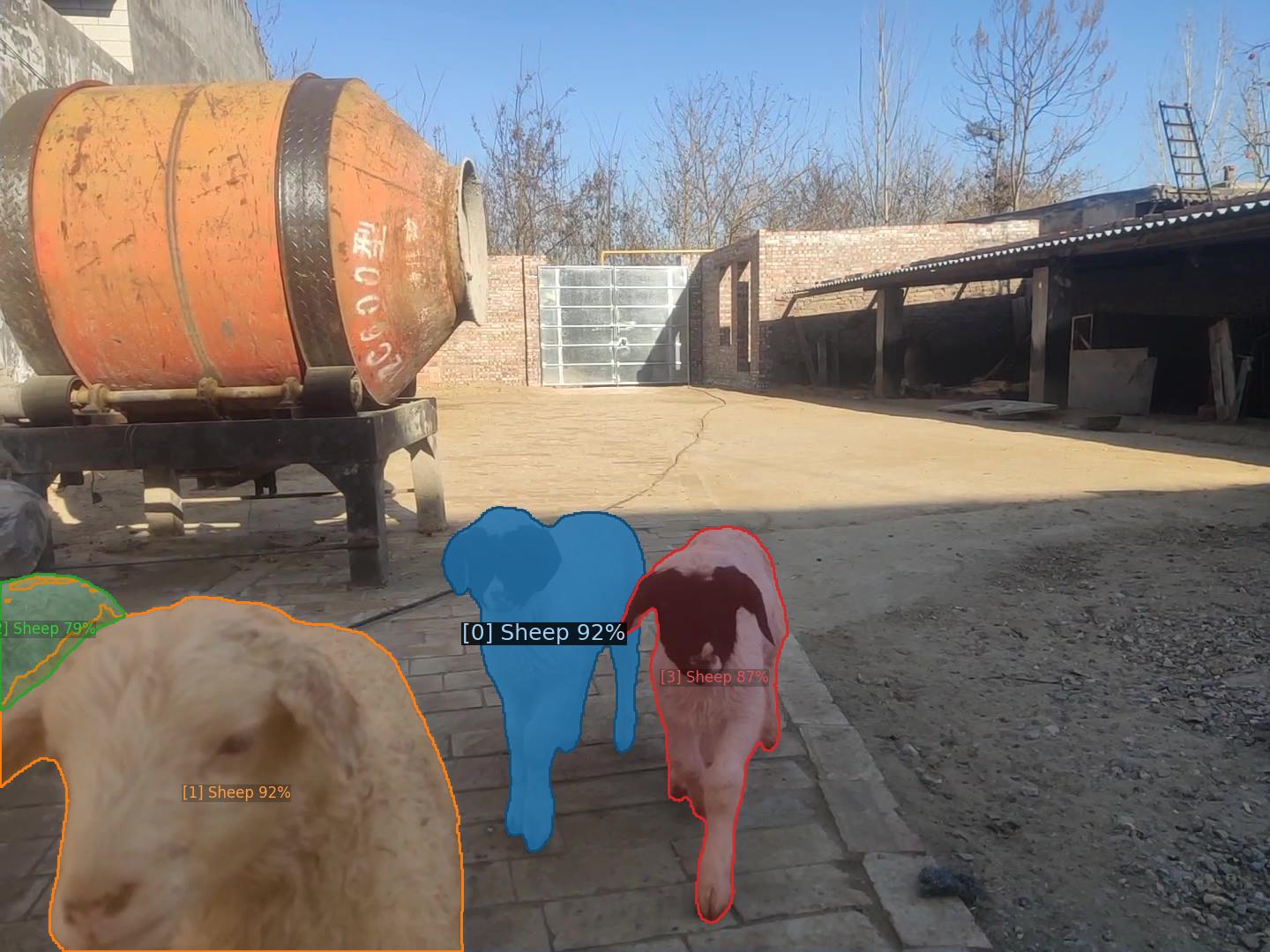}    & 
		\includegraphics[width=0.197\linewidth]{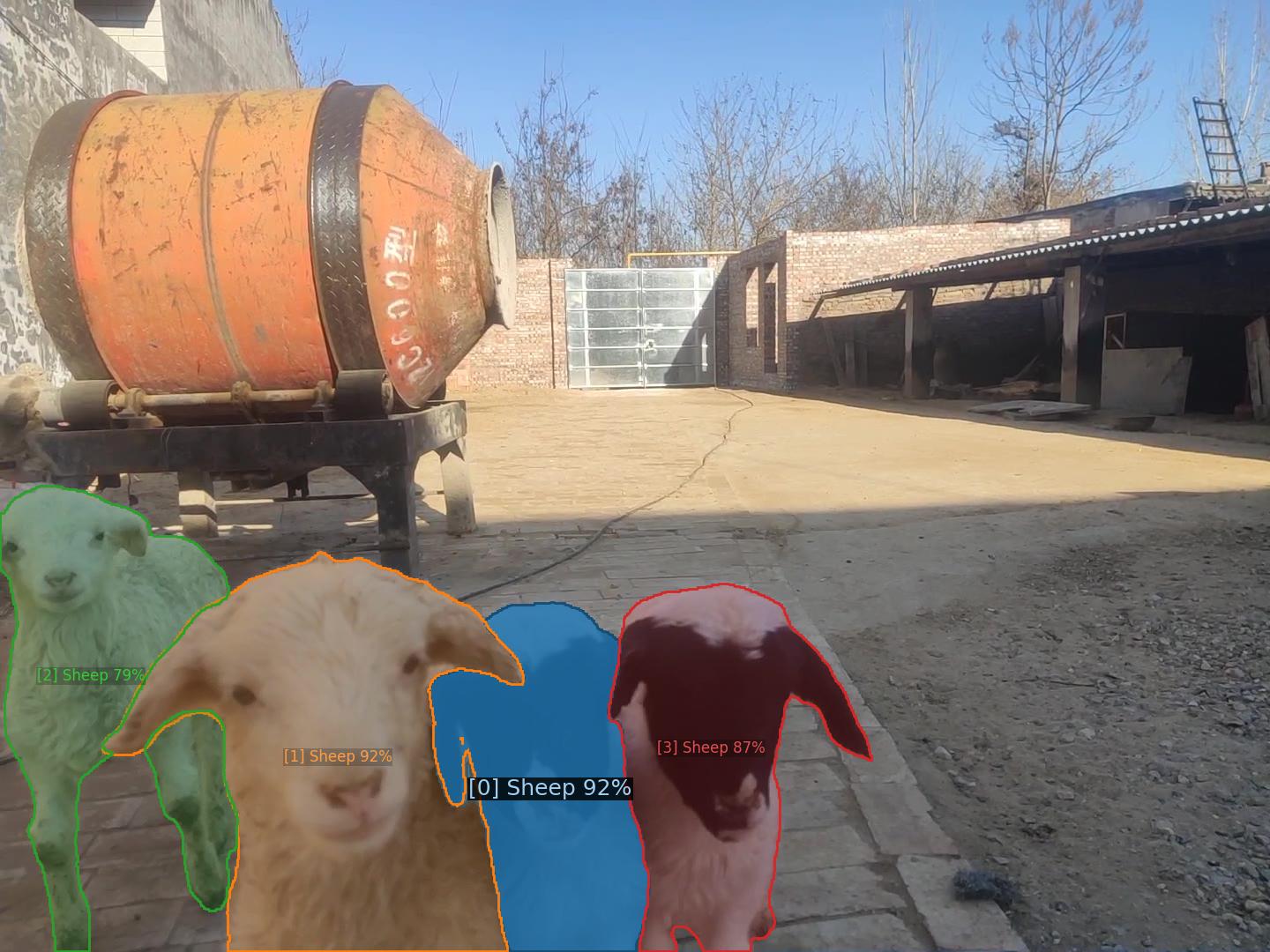}    & 
		\includegraphics[width=0.197\linewidth]{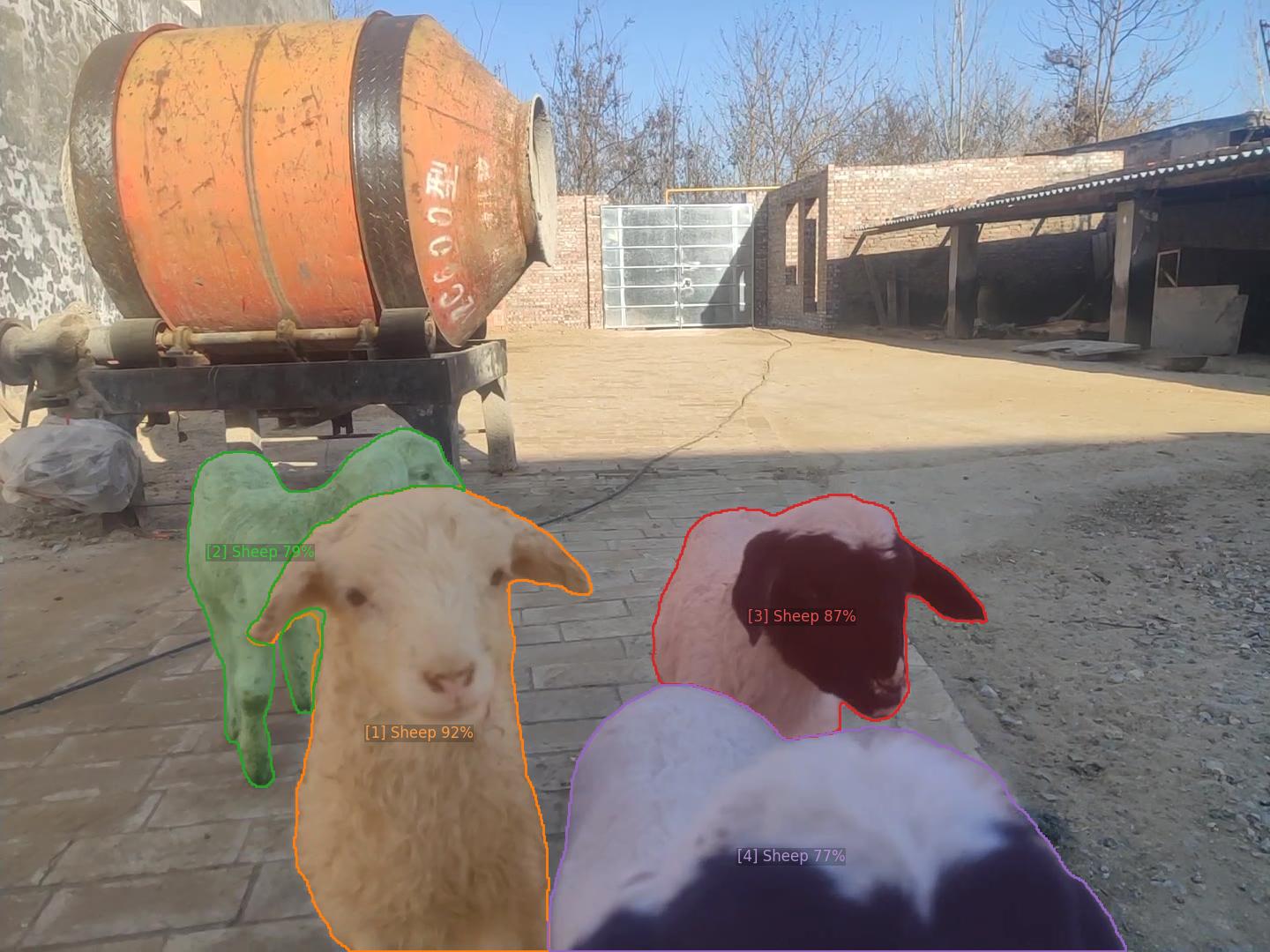}    &
		\includegraphics[width=0.197\linewidth]{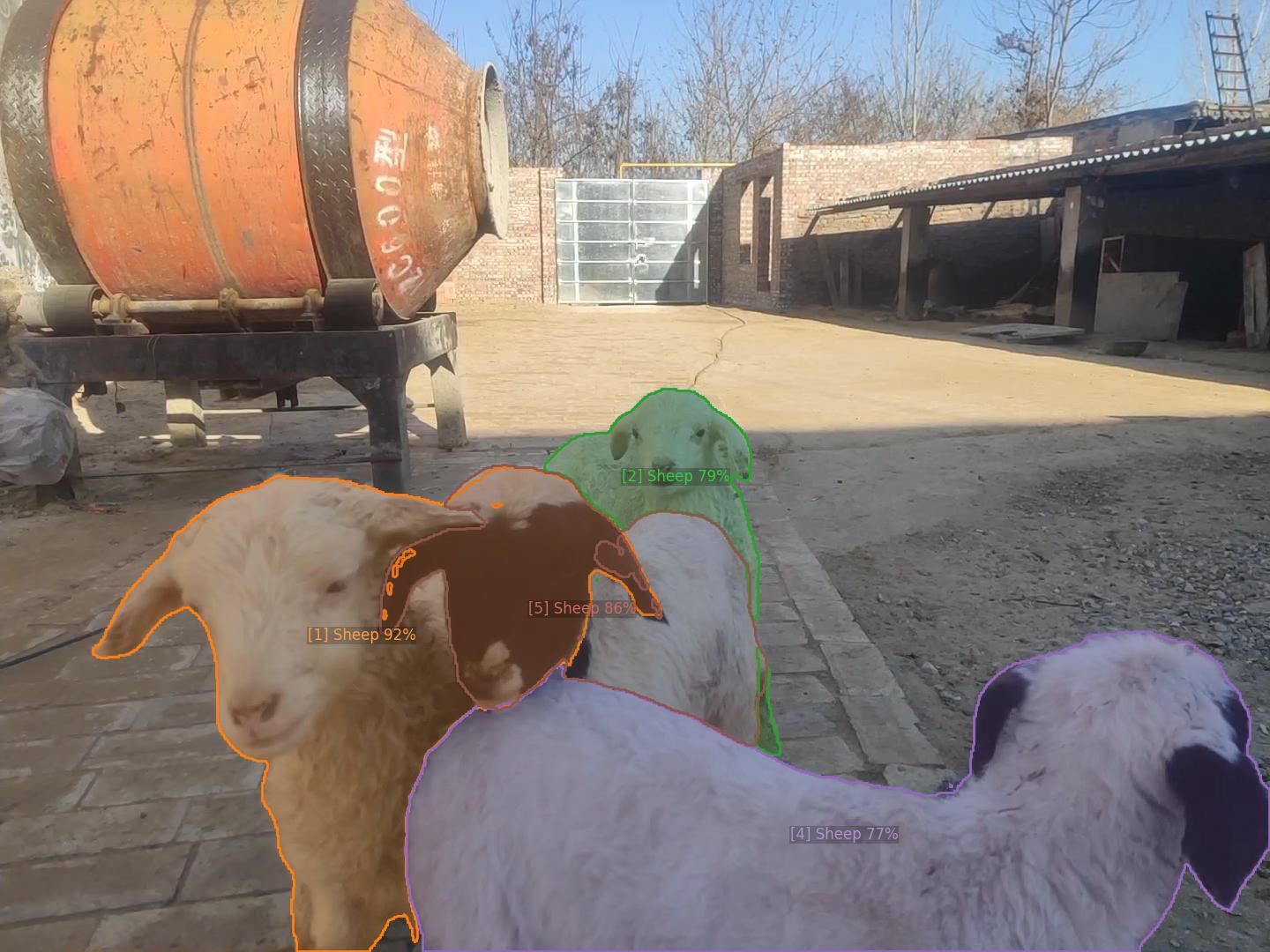} &
        \includegraphics[width=0.197\linewidth]{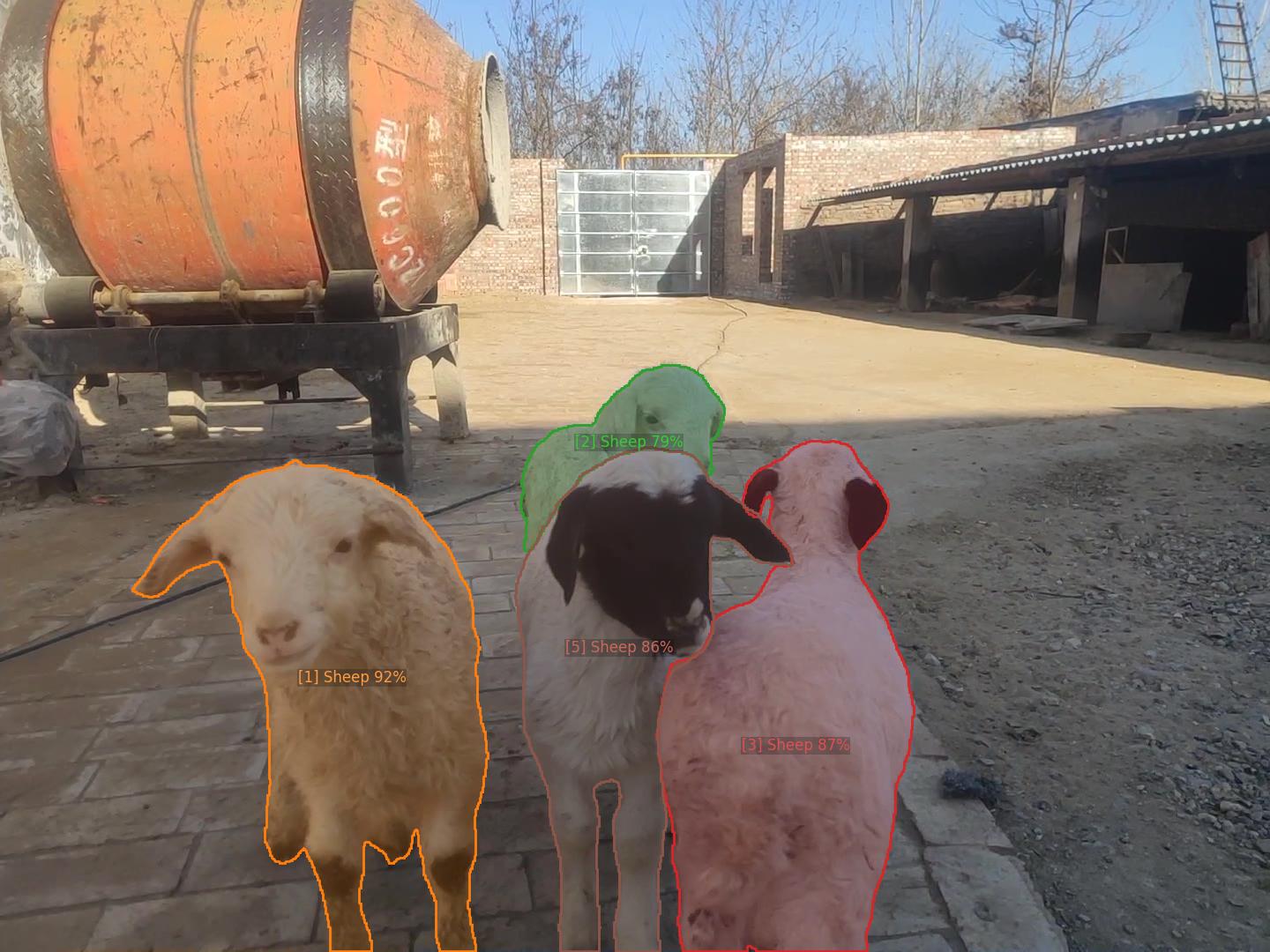}
		\\
		\includegraphics[width=0.197\linewidth]{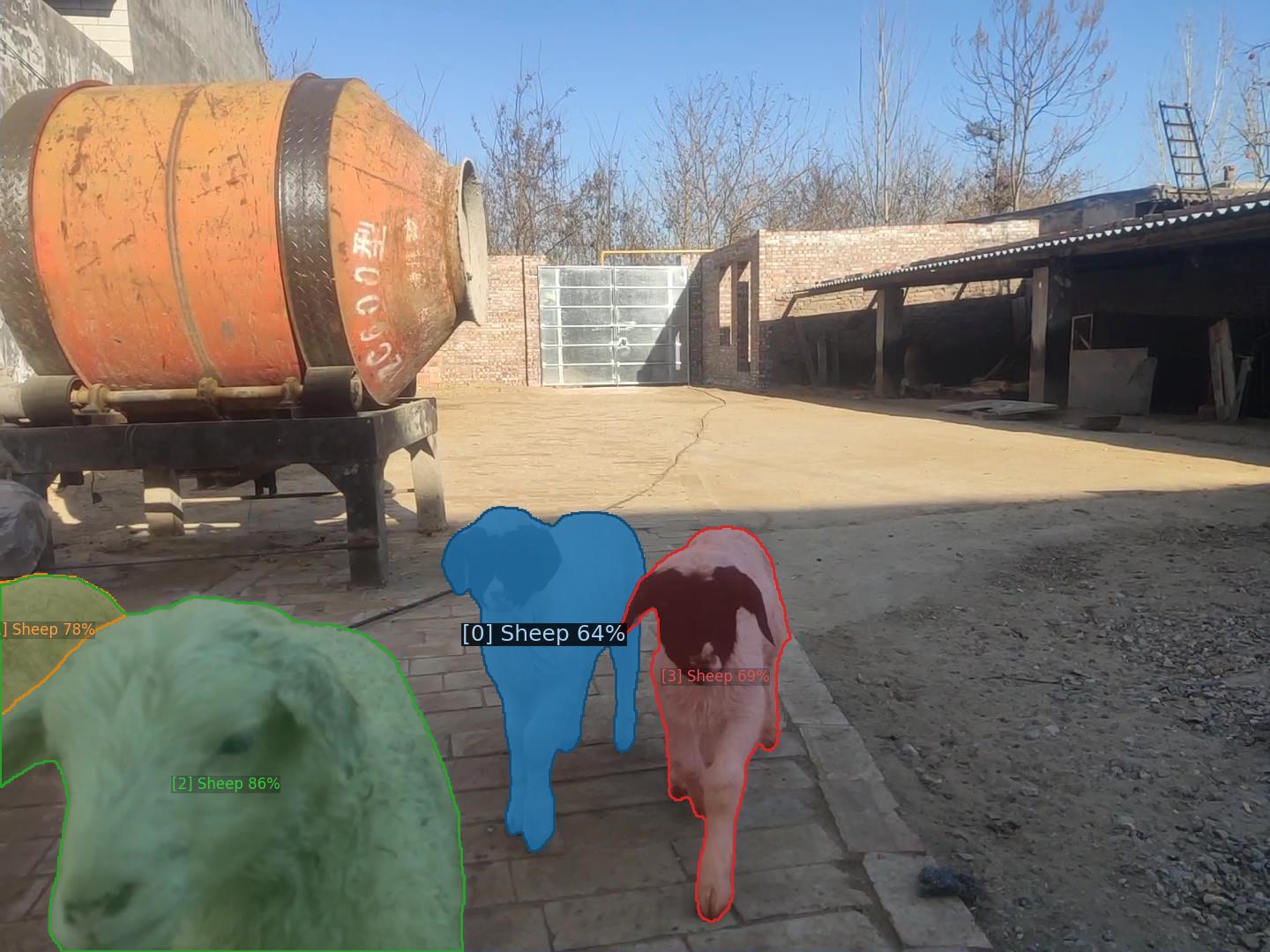}    & 
		\includegraphics[width=0.197\linewidth]{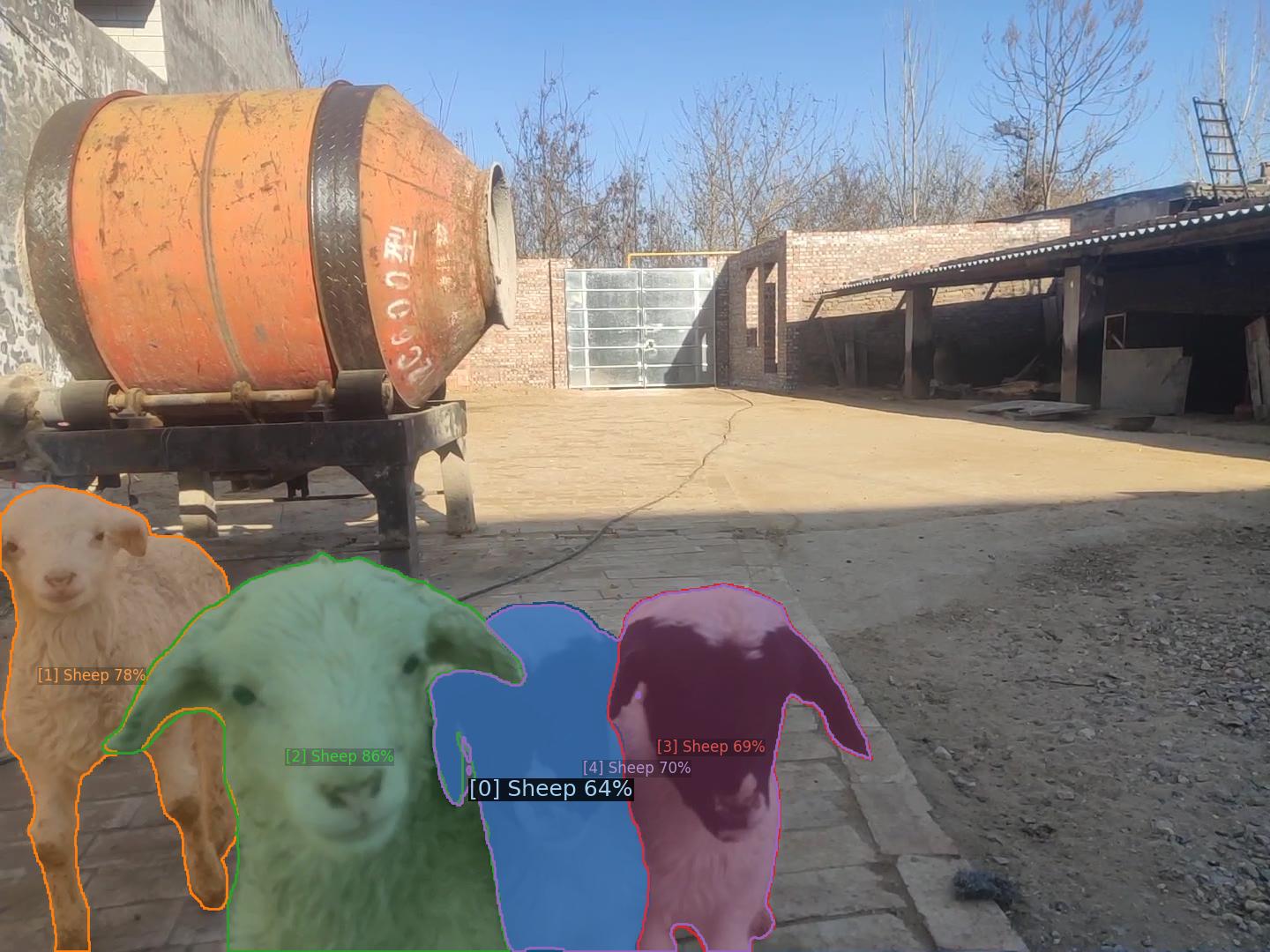}    & 
		\includegraphics[width=0.197\linewidth]{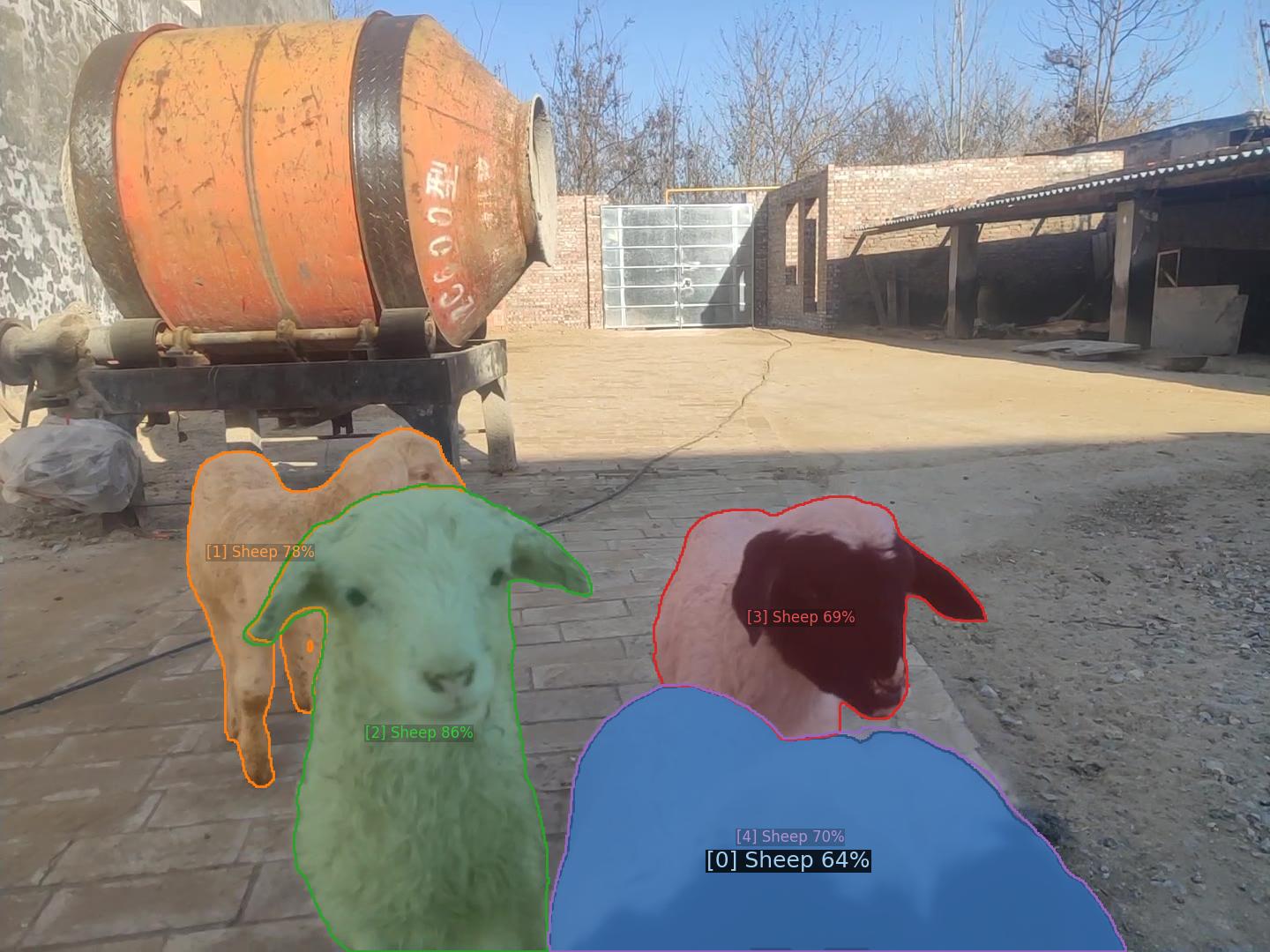}    &
		\includegraphics[width=0.197\linewidth]{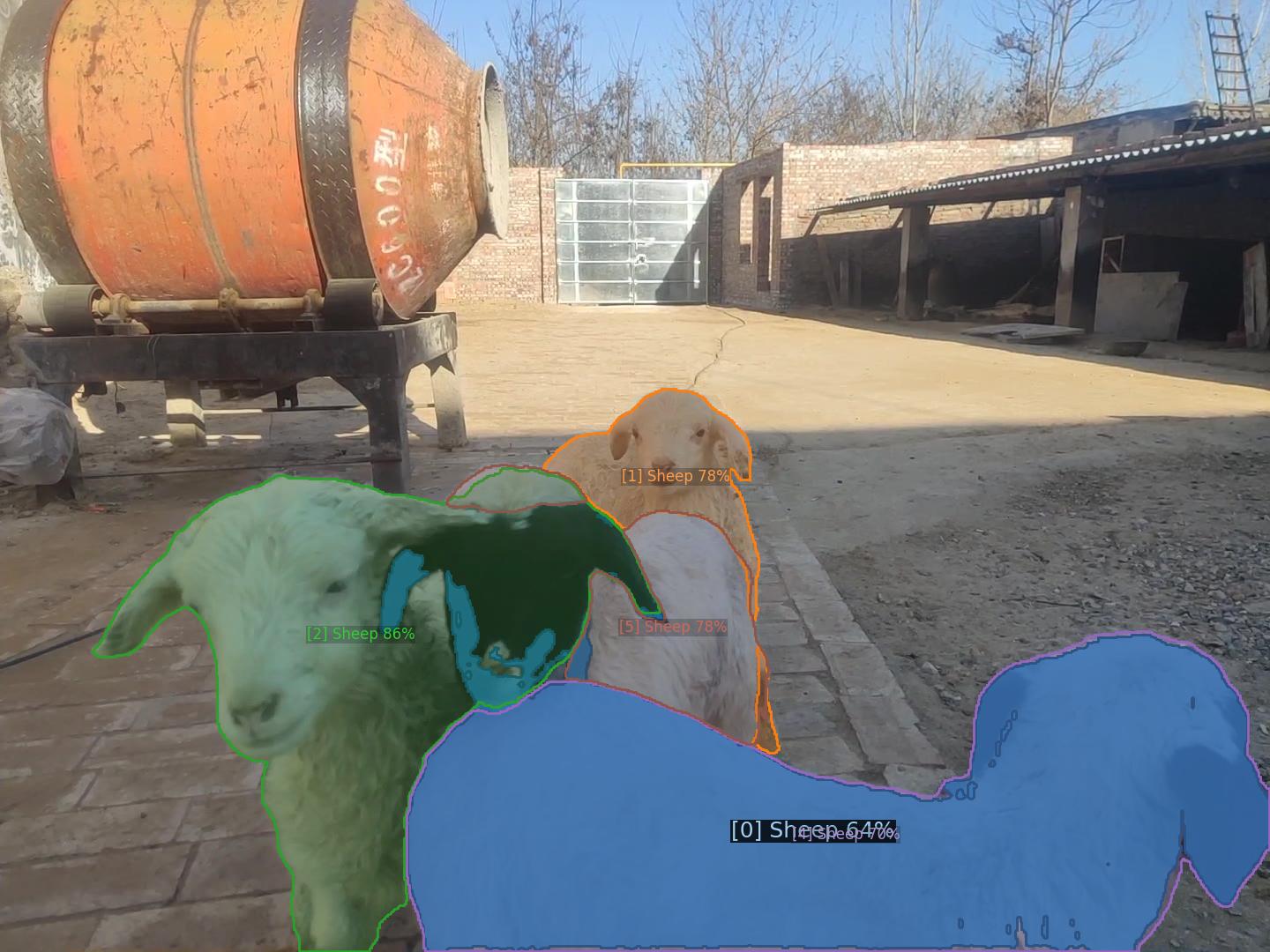} &
        \includegraphics[width=0.197\linewidth]{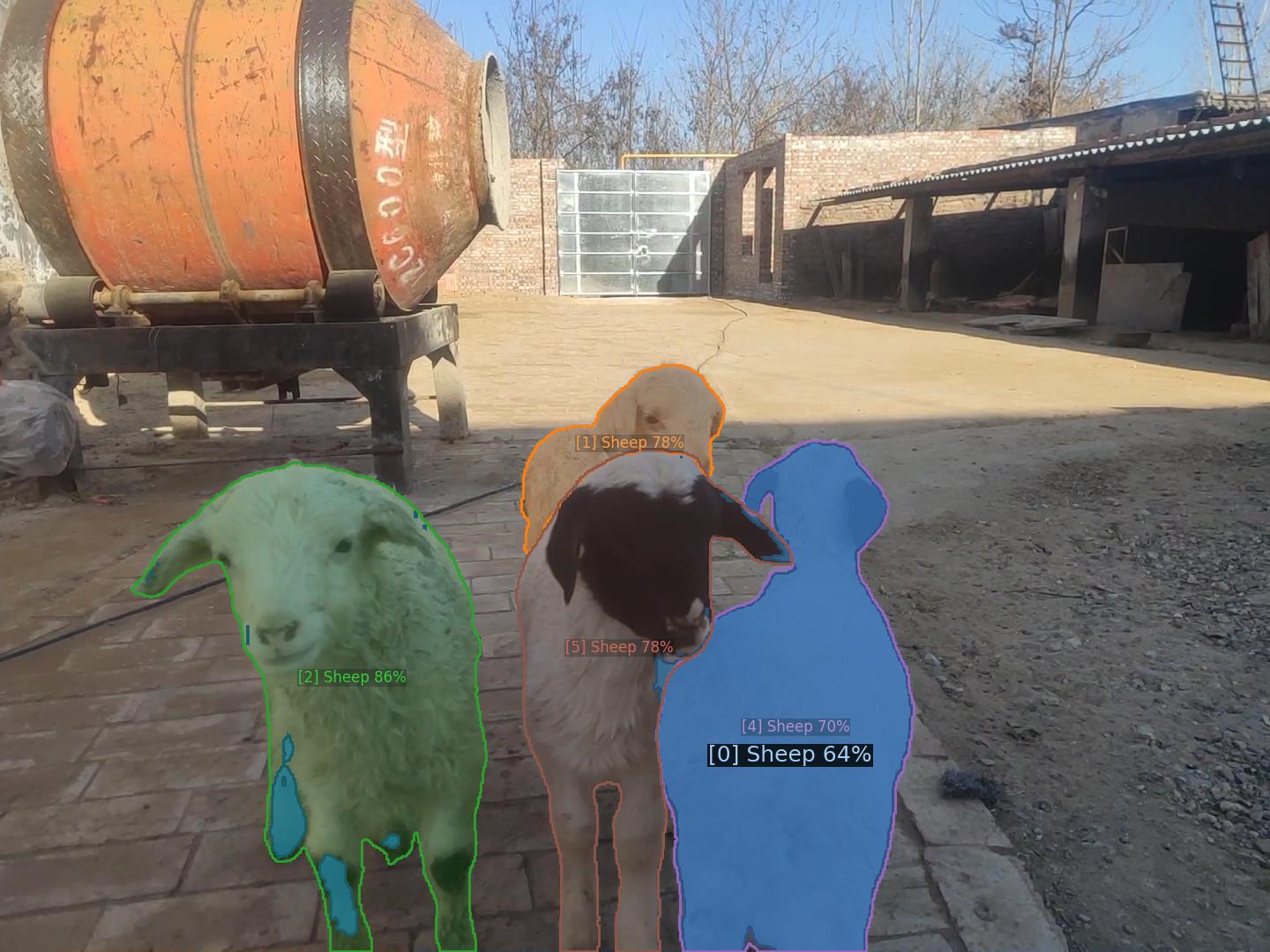}
		\end{tabular}

	\caption{Visual comparison of the mask predictions of selected objects \textbf{before} and \textbf{after} temporal refinement. Three selected videos from the OVIS dataset are visualized. For each video, the first row is \textbf{before} temporal refinement, and the second is \textbf{after} refinement. One can see that, in the first two videos, the masks for the highlighted duck and motorcycle are missing in the third frame before temporal refinement and are recovered after refinement. In the third video, the instance segmentations for the highlighted sheep are interrupted starting at the third frame before temporal refinement. Instance masks are more consistent after temporal refinement. }
	\label{fig:vis}
\end{figure*} \fi

\end{document}